%% file: main.tex
\documentclass[journal]{IEEEtran}

\IEEEoverridecommandlockouts
\usepackage{graphicx}
\def\BibTeX{{\rm B\kern-.05em{\sc i\kern-.025em b}\kern-.08em
    T\kern-.1667em\lower.7ex\hbox{E}\kern-.125emX}}
\usepackage[font=small,labelfont=bf,tableposition=top]{caption}

\usepackage{blindtext}
\usepackage{caption}

\let\oldtwocolumn\twocolumn
\renewcommand\twocolumn[1][]{%
    \oldtwocolumn[{#1}{
    \vskip-5ex
        \centering
        \includegraphics[width=1.0\textwidth]{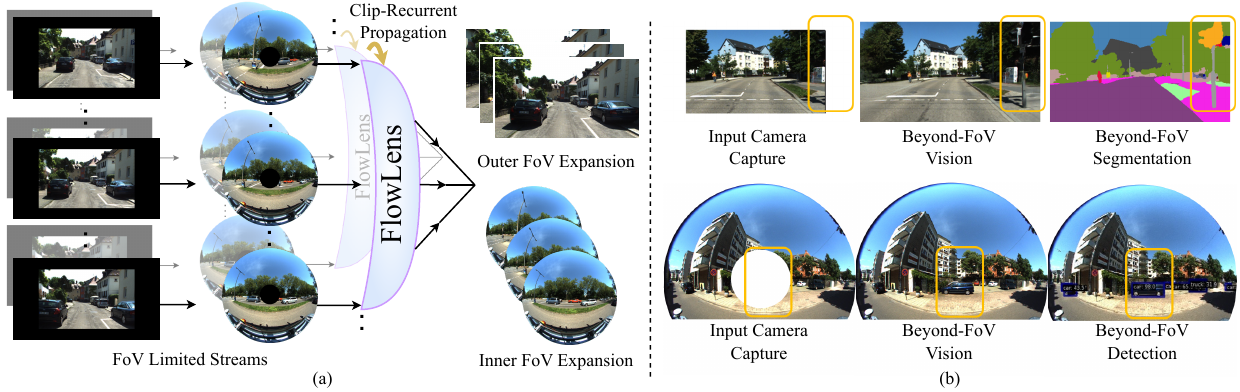}
        \captionof{figure} {\emph{Using the proposed FlowLens to enhance scene visibility and perception beyond vehicle-mounted cameras' FoV.} (a) FlowLens is capable of expanding the FoV of the camera through spatiotemporal feature propagation. (b) Comparing Input Camera Capture, Beyond-FoV Vision, and Perception: FlowLens enhances semantic segmentation of traffic lights and vehicle detection beyond the field of view, extending the perception boundary and enhancing autonomous driving system safety.}
        \label{fig:teaser}
    }]
}

\usepackage[table]{xcolor}
\definecolor{rblue}{rgb}{0,0.5,1}
\usepackage{hyperref}
\hypersetup{colorlinks=true,linkcolor=red,citecolor=rblue}

\usepackage{graphicx}

\usepackage{times}
\usepackage{epsfig}
\usepackage{graphicx}
\usepackage{amsmath}
\usepackage{amssymb}

\usepackage{lipsum}
\usepackage{stfloats}
\usepackage{multicol}
\usepackage{multirow}
\usepackage{bm}
\usepackage{etoolbox}

\usepackage{array}
\usepackage{tabulary}
\usepackage{paralist}
\usepackage{booktabs}
\usepackage{adjustbox}
\usepackage{xspace}
\usepackage[capitalize]{cleveref}
\crefname{section}{Sec.}{Secs.}
\Crefname{section}{Section}{Sections}
\Crefname{table}{Table}{Tables}
\crefname{table}{TABLE}{Tabs.}

\usepackage{makecell} 
\usepackage{arydshln} 
\usepackage{algorithm}
\usepackage{algorithmic}
\definecolor{commentcolor}{RGB}{110,154,155}   
\newcommand{\PyComment}[1]{\ttfamily\textcolor{commentcolor}{#1}}  
\newcommand{\PyCode}[1]{\ttfamily\textcolor{black}{#1}} 

\usepackage{xspace}
\makeatletter
\DeclareRobustCommand\onedot{\futurelet\@let@token\@onedot}
\def\@onedot{\ifx\@let@token.\else.\null\fi\xspace}

\def\ie{\emph{i.e}\onedot} 
 
\def\etc{\emph{etc}\onedot} 
 
\def\etal{\emph{et al}\onedot}
\makeatother

\usepackage{mathrsfs}
\usepackage{bbding}     

\usepackage[capitalize]{cleveref}
\crefname{section}{Sec.}{Secs.}
\Crefname{section}{Section}{Sections}
\Crefname{table}{Table}{Tables}
\crefname{table}{Tab.}{Tabs.}

\begin{document}
\title{Beyond the Field-of-View:\\Enhancing Scene Visibility and Perception\\with Clip-Recurrent Transformer}
\author{Hao Shi\IEEEauthorrefmark{1}, Qi Jiang\IEEEauthorrefmark{1}, Kailun Yang\IEEEauthorrefmark{2}, Xiaoting Yin, Ze Wang, and Kaiwei Wang\IEEEauthorrefmark{2}%
\thanks{This work was supported in part by the National Natural Science Foundation of China (NSFC) under Grant No. 12174341 and in part by Hangzhou SurImage Technology Company Ltd.}%
\thanks{H. Shi, Q. Jiang, X. Yin, Z. Wang, and K. Wang are with the State Key Laboratory of Extreme Photonics and Instrumentation and the National Engineering Research Center of Optical Instrumentation, Zhejiang University, Hangzhou 310027, China.}%
\thanks{K. Yang is with the School of Robotics and the National Engineering Research Center of Robot Visual Perception and Control Technology, Hunan University, Changsha 410082, China.}%
\thanks{H. Shi is also with Shanghai SUPREMIND Technology Company Ltd, Shanghai 201210, China.}%
\thanks{\IEEEauthorrefmark{1}Equal contribution.}%
\thanks{\IEEEauthorrefmark{2}Corresponding authors: Kaiwei Wang and Kailun Yang. (E-mail: wangkaiwei@zju.edu.cn, kailun.yang@hnu.edu.cn.)}%
}

\markboth{IEEE Transactions on Intelligent Vehicles, June~2024}%
{Shi \MakeLowercase{\textit{et al.}}: FlowLens}

\maketitle

\begin{abstract}
Vision sensors are widely applied in vehicles, robots, and roadside infrastructure. 
However, due to limitations in hardware cost and system size, camera Field-of-View (FoV) is often restricted and may not provide sufficient coverage.
Nevertheless, from a spatiotemporal perspective, it is possible to obtain information beyond the camera's physical FoV from past video streams.
In this paper, 
we propose the concept of online video inpainting for autonomous vehicles to expand the field of view, thereby enhancing scene visibility, perception, and system safety.
To achieve this, we introduce the \emph{FlowLens} architecture, which explicitly employs optical flow and implicitly incorporates a novel clip-recurrent transformer for feature propagation. FlowLens offers two key features:
1) FlowLens includes a newly designed Clip-Recurrent Hub with 3D-Decoupled Cross Attention (DDCA) to progressively process global information accumulated over time. 
2) It integrates a multi-branch Mix Fusion Feed Forward Network (MixF3N) to enhance the precise spatial flow of local features.
To facilitate training and evaluation, we derive the KITTI360 dataset with various FoV mask, which covers both outer- and inner FoV expansion scenarios. 
We also conduct both quantitative assessments and qualitative comparisons of beyond-FoV semantics and beyond-FoV object detection across different models.
We illustrate that employing FlowLens to reconstruct unseen scenes even enhances perception within the field of view by providing reliable semantic context. 
Extensive experiments and user studies involving offline and online video inpainting, as well as beyond-FoV perception tasks, demonstrate that FlowLens achieves state-of-the-art performance.
The source code and dataset are made publicly available at \url{https://github.com/MasterHow/FlowLens}.
\end{abstract}


\IEEEpeerreviewmaketitle

\section{Introduction}
\label{sec:intro}
\IEEEPARstart{V}{ision} sensors, including pinhole and spherical cameras, play a crucial role in acquiring visual information for intelligent vehicles and robots~\cite{hasan2021optical,kumar2023surround}. 
The capability to reconstruct an entire environment from partial observations is beneficial for mid-level tasks like object avoidance in mobile agents~\cite{kim2015rear}. Additionally, scene completion aids self-driving cars and robots in navigating and interacting with the real world~\cite{atapour2018comparative}.
However, due to constraints in hardware size and cost, the physical Field-of-View (FoV) of cameras may not always provide satisfactory coverage. 
Incomplete field-of-view observations, as depicted in Fig.~\ref{fig:teaser}(b), can lead to intelligent vehicles missing critical targets such as traffic lights or vehicles.
Nevertheless, when considering the spatiotemporal perspective, it becomes apparent that information beyond the current FoV can be readily obtained from past data streams. This raises a compelling question: \textbf{Can we extend our perception beyond the optical limits and sense the world outside the vehicle-mount camera's FoV?}

To address this question, we propose the integration of online video inpainting for autonomous vehicles, with the goal of expanding the field of view for camera sensors. This integration aims to empower these sensors to capture visual information beyond the physical limitations of optical systems, thereby enhancing scene visibility, perception, and system safety.
Our approach involves a hybrid propagation method for extending the camera's field of view. The fundamental idea is to utilize optical flow as a motion vector field to explicitly guide the propagation of temporal features. Simultaneously, we harness the capabilities of vision transformers~\cite{dosovitskiy2020image}, which can implicitly align features through self-attention mechanisms~\cite{shi2022rethinking}.
To put this idea into practice, we introduce \emph{FlowLens}, a novel \emph{Flow-guided Clip-Recurrent Transformer} designed for extending perception beyond the field of view. As depicted in Fig.~\ref{fig:teaser}(a), FlowLens leverages both past and current clips to provide a comprehensive view of the image. 

\begin{figure}[!t]
   \centering
   \includegraphics[width=1.0\linewidth]{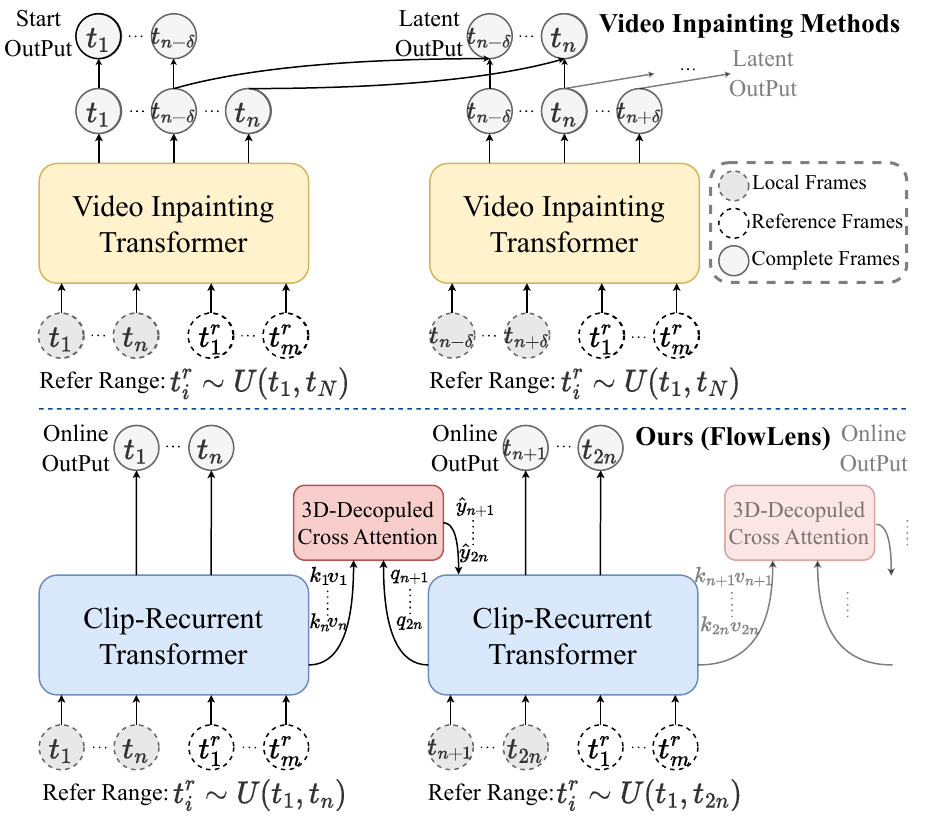}
   \caption{
   \emph{High-level logic comparison between FlowLens' Clip-Recurrent Transformer and the recent Video Inpainting Transformer (VI-Trans)~\cite{li2022towards,liu2021fuseformer}.} FlowLens provides immediate online output using only past streams and current clips. In contrast, VI-Trans generates delayed outputs dependent on future iterations. FlowLens' instant output capability makes it a more suitable choice for potential real-time applications, such as intelligent vehicle systems.
   }
   \label{fig:logic_compare}
\end{figure}

As a video editing technique, video inpainting~\cite{kim2019deep,xu2019deep} aims to fill in the missing part of the video in an offline mode.
Compared with the recent video inpainting transformer (VI-Trans)~\cite{li2022towards,liu2021fuseformer}, the main differences of FlowLens are three-fold (see~Fig.\ref{fig:teaser}(b-c)): 
(1)~\textbf{\emph{Online Output.}} VI-Trans relies on the future reference frame inputs and outputs are offline, whereas the output of FlowLens is immediate and online, which serve as critical prerequisites for real-world applications like autonomous driving.
(2)~\textbf{\emph{Past Reference Sampling.}} VI-Trans samples reference frames from the entire video, but when extending FoV we can only sample from past streams and cannot access future information.
(3)~\textbf{\emph{Clip-Recurrent Propagation.}} FlowLens can propagate the past clip features to the current iteration to fully exploit the potential of past reference frames.

Specifically, FlowLens operates on the current clip, but its ``queries'' additionally attend to the ``keys'' and ``values'' that are encoded and cached in the previous video clip.
This recurrent process allows the model to pass valuable encoded features from the past.
To achieve this, we introduce a \emph{Clip-Recurrent Hub} at the early stage of the transformer architecture.
Additionally, we propose a novel \emph{3D-Decoupled Cross Attention (DDCA)} mechanism to implement the query operation. By decoupling past 3D features in the spatiotemporal dimension, our model efficiently propagates relevant clip features for extending the FoV.
To further enhance FlowLens' ability to extract multi-scale local information, we introduce a new \emph{Mix Fusion Feed Forward Network (MixF3N)}. This network involves splitting into two depth-wise convolution branches with different kernel sizes after a soft composite operation, followed by a soft split operation to recover tokens.

Furthermore, to facilitate training and evaluation, we generate bidirectional FoV expansion masks using camera parameters for the KITTI360 dataset, which contains a total of $76,000$ frames of pinhole and spherical images. 
Pinhole cameras exhibit limitations primarily at the outer boundaries of the image plane, whereas spherical cameras with large FoV often face limitations related to the loss of central FoV (see Fig.~\ref{fig:teaser}(a)) introduced by reflective or catadioptric optics~\cite{gao2022review,zhang2020design,niu2007design}.
As a result, the derived dataset includes two subsets, providing masks for both the outer FoV extension of pinhole cameras and the inner FoV extension of spherical cameras, respectively.
We conducted benchmark tests on representative image inpainting models~\cite{wang2019wide,suvorov2022resolution,liu2022reduce} 
and video inpainting models~\cite{li2022towards,liu2021fuseformer,zeng2020learning} using the KITTI360 dataset. 
In addition, we perform quantitative and qualitative comparisons of the semantic segmentation and object detection results produced by FlowLens and other Video Inpainting Transformers (VI-Trans). The analysis demonstrated that FlowLens not only enhances semantic segmentation accuracy in previously unseen areas but can also improve segmentation within the original field of view. This is achieved by supplementing trusted context, ultimately enhancing the perception and overall system safety.
Experimental results and user studies on both offline and online video inpainting, as well as beyond-FoV perception tasks, reveal that FlowLens achieves state-of-the-art performance. 

In summary, we deliver the following contributions:
\begin{compactitem}
\item[(1)] 
We propose \emph{FlowLens}, a novel clip-recurrent transformer framework designed to enhance scene visibility and perception beyond the field of view in real-time, particularly valuable for applications in intelligent vehicles. 
\item[(2)] 
The newly introduced \emph{3D-Decoupled Cross Attention (DDCA)} and \emph{Mix Fusion Feed Forward Network (MixF3N)} are seamlessly integrated into the FlowLens architecture, further boosting its performance.
\item[(3)] 
Through extensive experiments and user studies, we demonstrate that FlowLens surpasses state-of-the-art video inpainting approaches in offline and online video inpainting, as well as beyond-FoV perception tasks. This advancement holds significant promise for improving the sensing capabilities of intelligent vehicle systems.
\end{compactitem}
\begin{figure*}[!t]
   \centering
   \includegraphics[width=1.0\linewidth]{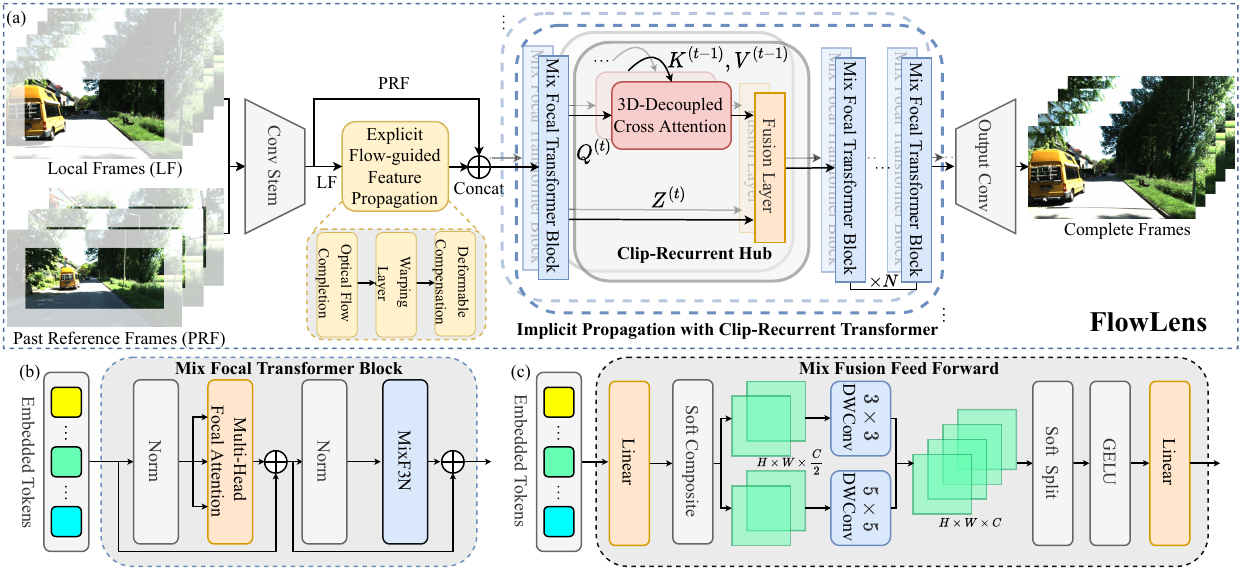}
   \caption{\emph{Illustrations of our proposed FlowLens.} (a) An overview. From left to right, it consists of 1) a convolution stem to extract shallow features, 2) an explicit flow-guided feature propagation module, 3) a clip-recurrent transformer to implicitly propagate features and fuse past information stream, 4) output convolution layers to restore the completed frames. (b)-(c) Our proposed Mix Focal Transformer Block and Mix Fusion Feed Forward Network (MixF3N).}
   \label{fig:overview}
\end{figure*}

\section{Related Work}
\label{sec:related_work}
\subsection{Beyond-FoV sensing.}
Researchers have been trying to obtain images with larger field-of-view for decades, with this exploration usually based on hardware expansion.
For pinhole cameras, people try to use multiple cameras for stitching~\cite{huang2014360,cowan2019360}.
On the other hand, spherical cameras naturally have the advantage of a large FoV. For example, a fisheye optical system can achieve a horizontal FoV of ${\sim} 180^\circ$~\cite{samy2015simplified,geng2017optical}, and a spherical camera that introduces reflective or catadioptric optical elements can further obtain a complete FoV of $360^\circ$ in the horizontal direction~\cite{zhang2020design,niu2007design}.
However, obtaining such a complete FoV in a single direction is often accompanied by a \textbf{central blind spot}~\cite{gao2022review}.
Beyond-FoV perception algorithms are also flourishing, including semantic segmentation~\cite{song2018im2pano3d,yang2019can,zhang2021transfer}, depth estimation~\cite{wang2020bifuse,zioulis2018omnidepth}, optical flow estimation~\cite{li2022deep,shi2022panoflow}, \etc.
Song~\etal~\cite{song2018im2pano3d} explore depth- and semantic estimation beyond the FoV for indoor layouts with a multi-stream convolutional neural network (CNN). Boundless~\cite{teterwak2019boundless} employs an improved image inpainting generative adversarial network (GAN)~\cite{yu2019free} for one-way extension and panoramic synthesis.
Cylin-Painting~\cite{liao2023cylin} combines the image inpainting and outpainting techniques to achieve a visually pleasant cylinder from a partial view.
The closest works to ours are FisheyeEX~\cite{liao2022fisheyeex} and FishDreamer~\cite{shi2023fishdreamer}, which attempt to expand the FoV of a fisheye camera outwards by single image outpainting techniques~\cite{sabini2018painting}.
Instead, we introduce a novel clip-recurrent transformer to propagate and generate more temporal coherence results for both the outer expansion of pinhole cameras and the inner expansion of spherical cameras.

\subsection{Inpainting.}
Inpainting techniques aim to sample and fill known textures with plausible imagery into holes of an image for removing watermarks or unwanted image content~\cite{bertalmio2000image,bertalmio2003simultaneous,barnes2009patchmatch}. With the development of deep learning, many works~\cite{pathak2016context,iizuka2017globally,liu2018image,yu2019free,yi2020contextual,li2022misf} have explored image inpainting techniques based on GAN~\cite{goodfellow2020generative}. Recently, LaMa~\cite{suvorov2022resolution} explores the single-stage image inpainting based on large mask training and fast Fourier convolutions.
PUT~\cite{liu2022reduce} introduces a transformer model based on the patch auto-encoder.
In addition to image inpainting, some efforts~\cite{kim2021painting_edge_guided,cheng2022inout,kong2022image_adaptive_hint} have been made to expand the image outward (\ie outpainting).
The representative work is SRN~\cite{wang2019wide}, which introduces a context normalization module and a spatial variant loss 
to predict the texture outward.
However, these models usually focus on estimating the outer side of the image and cannot address the inner missing regions, and the expansion rate is bound to the model structure, especially for transformer-based models~\cite{gao2022generalised,yao2022outpainting}. 
Unlike image inpainting, video inpainting can additionally model temporal dependencies~\cite{newson2014video,strobel2014flow,huang2016temporally}.
From the perspective of feature propagation, learning-based video inpainting methods can be divided into two categories: explicit propagation models and implicit propagation models.
The explicit propagation models integrate optical flow as a strong prior into the framework~\cite{xu2019deep,gao2020flow,kang2022ecfvi,zhang2022fgt,gu2023flow}, among them the representative work is E2FGVI~\cite{li2022towards}, which builds an end-to-end flow-guided feature propagation structure for video inpainting. {FGDVI~\cite{gu2023flow} integrates optical flow as a motion prior into the diffusion model to improve spatiotemporal continuity.}
In addition, ProPainter~\cite{zhou2023propainter} first uses RAFT~\cite{teed2020raft} to calculate the complementary optical flow of the image pair to perform pixel warping, and then exploit the flow again to perform feature warping, further improving the accuracy.
Implicit feature propagation relies on the 3D convolution~\cite{kim2019deep,chang2019free,lee2019copy,wang2019video} or the self-attention mechanism of transformers~\cite{liu2021fuseformer,zeng2020learning,yu2023deficiency} to regroup and assign known inter-frame textures. {DMT~\cite{yu2023deficiency} distills knowledge from an image inpainting model to enhance the generation of visually plausible content.}
Different from existing methods, FlowLens benefits from both explicit flow-guided propagation and implicit transformer propagation.
Besides, our approach can additionally retrieve relevant values from past iterations with the clip-recurrent propagation to boost the performance.

\subsection{Deep Generative Models.}
Recently, deep generative models have gained considerable attention, including GAN-based approaches~\cite{goodfellow2014generative,clark2019adversarial}, autoregressive transformers~\cite{brown2020language,babaeizadeh2021fitvid,liu2024sora}, non-autoregressive transformers~\cite{chang2022maskgit,gupta2022maskvit}, and, of course, diffusion methods~\cite{austin2021structured,sohl2015deep}.
Among these appealing methods, we employ video GAN~\cite{clark2019adversarial} as the fundamental paradigm for achieving beyond-FoV. Compared to the video diffusion~\cite{ho2022imagen,gu2023flow,wu2024towards} model, GAN can obtain satisfactory results in a single forward pass instead of multiple diffusion processes, which is crucial for potential real-time applications. Furthermore, compared to the VQ-VAE~\cite{van2017neural,razavi2019generating} method, GAN does not require multiple training stages or rely on large-scale data for pre-training of the latent representation learning. Consequently, FlowLens adopts the popular video GAN in the video editing domain as the baseline, primarily due to the trade-off between sampling speed, data consumption, and generation quality.

\section{Methodology}
\noindent\textbf{Preliminary.}
Given a FoV-limited video sequence $\mathbf{X}^t=\{X^t \in \mathbb{R}^{H \times W \times 3} |t=1...T \}$
and a corresponding binary mask $M \in \mathbb{R}^{H \times W \times 1}$ representing the missing FoV, whose values are either $0$ denoting the original image plane or $1$ referring to the regions that require filling.
Note that $M$ can be a sequence of different FoVs during training, but it keeps the same during testing considering the actual limitation of FoV-limited cameras.
The goal of our FoV expansion task is to propagate plausible and spatiotemporally coherent content from $\mathbf{X}^t$ to the complete frames $\mathbf{\hat{Y}}^t=\{\hat{Y}^t \in \mathbb{R}^{H \times W \times 3} |t=1...T \}$ with a larger FoV.

\noindent\textbf{Overview.}
Fig.~\ref{fig:overview} shows the entire pipeline of the proposed \emph{FlowLens} for FoV expansion.
We first use a convolutional stem 
to encode the input Local Frames (LF) and Past Reference Frames (PRF) sampled from $\mathbf{X}^t$.
Then, the LF features are fed into the \emph{Explicit Flow-guided Feature Propagation} module (Sec.~\ref{sec:explicit_propagation}) to complete the feature under the motion prior.
Next, the \emph{Clip-Recurrent Transformer} (Sec.~\ref{sec:clip_recurrent_transformer}), with our proposed \emph{Mix Fusion Feed Forward Network (MixF3N)}, queries and implicitly aligns spatiotemporally related features in the PRF with the LF, and retrieves the coherence values from the last iteration via the \emph{Clip-Recurrent Hub} that is equipped with our \emph{3D-Decoupled Cross Attention (DDCA)}.
Finally, we adopt the output convolutional layers to up-sample the complete features back to the original scale and reconstruct the FoV-expanded sequence $\mathbf{\hat{Y}}^t$.

\subsection{Explicit Propagation with Optical Flow}
\label{sec:explicit_propagation}
\begin{figure}[!t]
  \centering
  \includegraphics[width=0.9\linewidth]{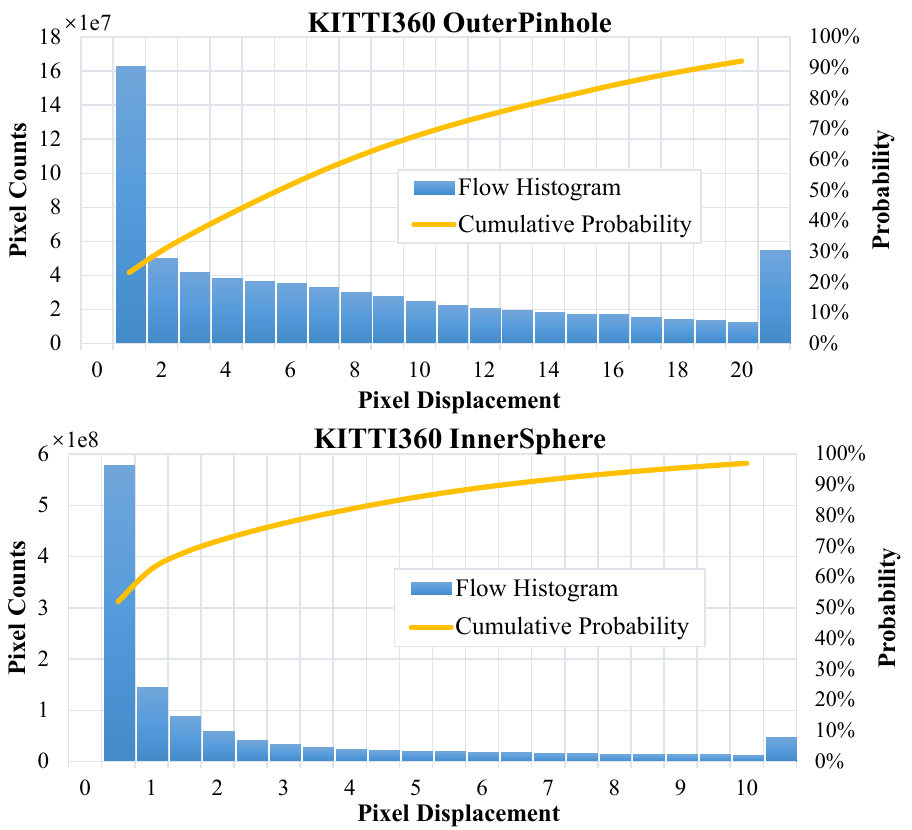} %
%
  \caption{\emph{The histogram of the movement for the KITTI360.} The distribution of pixel displacement exhibits a long-tailed feature and contains large motion, especially for the pinhole camera.}
  \label{fig:kitti_flow}
\end{figure}
Our proposed \emph{Explicit Flow-guided Feature Propagation} module consists of three stages (see Fig.~\ref{fig:overview}(a)): optical flow completion, warping layer, and deformable compensation.
As previously shown in~\cite{xu2019deep}, it is simpler to complete the corrupted optical flow than directly estimating the missing pixels.
Therefore, we first estimate the beyond-FoV optical flow for two adjacent FoV-limited frames $X^i$ and $X^j$ by a flow completion network $\mathcal{F}$:
%
\begin{equation}
\label{equ:flow_complete}
    \begin{aligned}
    \hat{\mathbf{V}}_{i \rightarrow j}=\mathcal{F}(d_{4}(X^{i}),d_{4}(X^{j})),
    \end{aligned}
\end{equation}
where $d_4(\cdot)$ denotes downsampling by $\frac{1}{4} \times$.
As shown in Fig.~\ref{fig:kitti_flow}, the distribution of pixel movement on KITTI360 exhibits a long-tailed feature and contains large displacements, especially for the pinhole camera outer estimation.

According to the observed motion distribution, we consider that the accurate flow prior is important for feature propagation, and thus we propose to incorporate a strong flow completion network that generates a $6-$level shared feature pyramid 
and makes predictions from level $6$ to $2$
in a coarse-to-fine fashion to output a high-quality flow field.

Next, we exploit 
the motion prior in order to warp the input local frames in the feature space to mitigate the inaccurate flow estimation and 
frame-level occlusion.
Specifically, given the feature maps $f_{i}, f_{j} \in \mathbb{R}^{\frac{H}{4} \times \frac{W}{4} \times C}$ of local frames, we have:
%
\begin{equation}
\label{equ:flow_warp}
    \begin{aligned}
    \tilde{f}_{i}(\mathbf{x})=f_{j}(\mathbf{x}+\hat{\mathbf{V}}_{i \rightarrow j}(\mathbf{x})),
    \end{aligned}
\end{equation}
where $\tilde{f}_{i}$ is the first-order propagation feature and $\mathbf{x}$ is the pixel index. 
In explicit feature propagation, an important challenge lies in aligning multiple frames. Advanced video processing models, such as state-of-the-art video super-resolution transformers~\cite{lin2021fdan,liang2022vrt,chan2022basicvsr++} and frame interpolation networks~\cite{lee2020adacof,ding2021cdfi} are generally equipped with well-designed alignment modules combining optical flow and deformable convolution networks (DCN). FlowLens is no exception, as we introduce DCN
after the warping layer to sample from diverse spatial locations, the error accumulation caused by flow-guided feature propagation can be further compensated.

More precisely, we compute the offset $\mathbf{o}_{i \rightarrow j}$ and the modulation 
weight $m_{i \rightarrow j}$ of modulated deformable convolution~\cite{zhu2019deformable} based on the flow prediction:
%
\begin{equation}
\label{equ:dcn_compensate_01}
    \begin{aligned}
    & \mathbf{o}_{i \rightarrow j}=\hat{\mathbf{V}}_{i \rightarrow j} + r_{max} \cdot tanh(\mathcal{C}_{off}(\tilde{f}_{i}, f_{i}, \hat{\mathbf{V}}_{i \rightarrow j})),\\
    & m_{i \rightarrow j}=\sigma( \mathcal{C}_{mod}(\tilde{f}_{i}, f_{i}, \hat{\mathbf{V}}_{i \rightarrow j}))),
    \end{aligned}
\end{equation}
where $\mathcal{C}_{off}$ and $\mathcal{C}_{mod}$ are sets of convolutional layers and $\sigma$ represents the Sigmoid activation function.
We introduce $r_{max}$ as the max compensate residue magnification, which is set to $10$ in all experiments.
The DCN is then applied as:
%
\begin{equation}
\label{equ:dcn_compensate_02}
    \begin{aligned}
    \hat{f}_{i} = \mathcal{C}_{prop}(f_{i}, \mathcal{D}(f_{j} | \mathbf{o}_{i \rightarrow j}, m_{i \rightarrow j})),
    \end{aligned}
\end{equation}
where $\hat{f}_{i}$ is the second-order propagation feature, $\mathcal{C}_{prop}$ is a two-layer stacked 3${\times}$3 convolution, and $\mathcal{D}$ denotes the DCN.
The above propagation is performed between adjacent local frames in a bidirectional manner.
We finally apply $\mathcal{C}_{fuse}$ as a $1 {\times} 1$ convolutional layer to adaptively fuse the forward- and backward propagation feature $\hat{f}^{f}_{i},\hat{f}^{b}_{i}$:
%
\begin{equation}
\label{equ:dcn_compensate_03}
    \begin{aligned}
    \hat{F}_{i} = \mathcal{C}_{fuse}(\hat{f}^{f}_{i}, \hat{f}^{b}_{i}).
    \end{aligned}
\end{equation}

\subsection{Clip-Recurrent Transformer}
\label{sec:clip_recurrent_transformer}
For beyond-FoV estimation, it is not sufficient to only rely on the current clip. Since future videos are not available, we must consider how to further explore past information streams. To this end, we propose a novel \emph{Clip-Recurrent Transformer} framework, with a \emph{Clip-Recurrent Hub} enhanced transformer architecture and use \emph{3D-Decoupled Cross Attention (DDCA)} to query the spatiotemporal coherent features from the previous iteration. Another challenge comes from the local finesse.
As previously shown in MiT~\cite{xie2021segformer}, convolutions could leak local information in a direct way while maintaining performance.
Therefore, we introduce the \emph{Mix Fusion Feed Forward Network (MixF3N)} to further facilitate the local features flow between soft split tokens in a multi-scale manner for performing the FoV expansion.

As shown in Fig.~\ref{fig:overview}(a), the Clip-Recurrent Transformer consists of a Clip-Recurrent Hub and $N$ Mix Focal Transformer blocks.
The Mix Focal Transformer blocks are the same as Temporal Focal Transformer~\cite{li2022towards,yang2021focal} block except that the fusion feed-forward network is replaced with our proposed MixF3N.
Suppose $\hat{F}_{lf} \in \mathbb{R}^{T_{lf} \times \frac{H}{4} \times \frac{W}{4} \times C}$ and $F_{pf} \in \mathbb{R}^{T_{pf} \times \frac{H}{4} \times \frac{W}{4} \times C}$ are encoded local- and past reference frames, respectively, we use soft split~\cite{liu2021fuseformer} to embed them into overlapped patches $X \in \mathbb{R}^{(T_{lf}+T_{pf}) \times W_{h} \times W_{w} \times C_{e}}$:
%
\begin{equation}
\label{equ:soft_split}
    \begin{aligned}
    X = {\rm SS}(\hat{F}_{lf} \oplus F_{pf}),
    \end{aligned}
\end{equation}
where ${\rm SS}(\cdot)$ denotes the soft split operation. $T_{lf}$ and $T_{pf}$ are the time dimension of LF and PRF, and $W_{h}$ and $W_{w}$ are the spatial dimension of embedded tokens. $\oplus$ denotes the feature concatenation.

Then, $X$ is linearly projected to queries $Q$, keys $K$, and values $V$ for computing the focal attention and MixF3N to obtain the output tensor $Z \in \mathbb{R}^{(T_{lf}+T_{pf}) \times W_{h} \times W_{w} \times C_{out}}$:
%
\begin{equation}
\label{equ:block}
    \begin{aligned}
    & Q, K, V = \mathcal{P}_{qkv}({\rm LN_{1}}(X)), \\
    & Z' = {\rm MHFA}(Q, K, V) + X, \\
    & Z = {\rm MixF3N}({\rm LN_{2}}(Z')) + Z',
    \end{aligned}
\end{equation}
where ${\rm LN}$ and ${\rm MHFA}$ denote the layer normalization and multi-head focal attention, respectively, $\mathcal{P}_{qkv}$ is the linear projection layer, and the key difference of our Mix Focal Transformer compared with the previous VI-Trans~\cite{li2022towards} lies in the newly-designed Mix Fusion Feed Forward Network (MixF3N). We omit the time dimension for simplicity.

\begin{algorithm}[t]
    \caption{Pseudo code of FlowLens Clip-Recurrent Hub in a PyTorch-like style.}\label{alg:recurrent}
    \input{codes/recurrent-code}

\end{algorithm}

\noindent\textbf{Clip-Recurrent Hub.}
To realize the progressive propagation of spatiotemporal correlation features along the iterations, we implement Clip-Recurrent Hub as the crucial transit center for information exchange.
Algorithm~\ref{alg:recurrent} presents the pseudo-code in a PyTorch-like style for our design.
Concretely, we first cache the keys and values  $\mathbf{K}^{t}_{i},\mathbf{V}^{t}_{i}=\{K^{t}_{i},V^{t}_{i} \in \mathbb{R}^{W_{h} \times W_{w} \times C_{e}} |t \in T_{lf} \cup T_{pf} \}$ of the $i$-th iteration:
%
\begin{equation}
\label{equ:cache_01}
    \begin{aligned}
    \bar{\mathbf{K}}^{t}_{i}, \bar{\mathbf{V}}^{t}_{i} := {\rm SG}(\mathbf{K}^{t}_{i}, \mathbf{V}^{t}_{i}),
    \end{aligned}
\end{equation}
where ${\rm SG}(\cdot)$ is the stop gradient operator to avoid past backward propagation. $\bar{\mathbf{K}}^{t}_{i}$ and $\bar{\mathbf{V}}^{t}_{i}$ denote the cached clip keys and values.
Whenever the camera is initiated (\ie $i=0$), $\bar{\mathbf{K}}^{t}_{0}$ and $\bar{\mathbf{V}}^{t}_{0}$ are associated with the same as $\mathbf{K}^{t}_{1}$ and $\mathbf{V}^{t}_{1}$, and it is iteratively updated at further time stamp $i>0$. 
Then, we introduce the 3D-Decoupled Cross Attention and query from the clip buffer at the next iteration:
%
\begin{equation}
\label{equ:cache_02}
    \begin{aligned}
    & \mathbf{\bar{Z}}'_{i+1} = {\rm DDCA}(\mathbf{Q}_{i+1}, \mathcal{P}_{kv}(\bar{\mathbf{K}}_{i}, \bar{\mathbf{V}}_{i})), \\
    & \mathbf{\hat{Z}}'_{i+1} = \mathbf{Z}'_{i+1} + \mathcal{P}_{fuse}(\mathbf{\bar{Z}}'_{i+1} \oplus \mathbf{Z}'_{i+1}), 
    \end{aligned}
\end{equation}
where ${\rm DDCA}$ denotes our proposed 3D-Decoupled Cross Attention, which will be described in detail below.
$\mathcal{P}_{kv}$ and $\mathcal{P}_{fuse}$ are linear projections for cache updating and token fusion, respectively.
$\mathbf{\bar{Z}}'_{i+1}$ is the spatiotemporal coherent feature queried from the previous clip, whereas $\mathbf{\hat{Z}}'_{i+1}$ is the clip-recurrent enhanced feature of the $(i+1)$-th iteration. 

The Clip-Recurrent Hub constitutes a core design of FlowLens.
Given an arbitrary frame with a missing FoV, FlowLens can either exploit the global information accumulated in the 
%
temporal dimension with the help of Clip-Recurrent Hub, or extract fine-grained local information between adjacent frames by the Mix Focal Transformer, which is essentially different from previous VI-Trans methods~\cite{li2022towards,liu2021fuseformer,zeng2020learning} that only extract features from local windows and limited reference frames.

\noindent\textbf{3D-Decoupled Cross Attention.}
An intuitive idea is to use vanilla multi-head self-attention (MSA)~\cite{dosovitskiy2020image} directly for the cross-attention implementation.
However, a recent study~\cite{park2022vision} on the nature of ViT suggests that the purely global receptive field of MSA may introduce unnecessary degrees of freedom and thus lack focus on locally relevant features.
Therefore, we propose 3D-Decoupled Cross Attention (DDCA) to search from both local and non-local spatiotemporal neighborhoods.
Specifically, DDCA first performs vanilla attention along the time axis:
%
\begin{equation}
\label{equ:time_att_01}
    \begin{aligned}
    Z_{t} := {\rm Attn}(Q_{t}, K_{t}, V_{t}) = {\rm Softmax}(\frac{Q_{t} K_{t}^{\top}V_{t}}{\sqrt{d}}),
    \end{aligned}
\end{equation}
where $Q_{t},K_{t},V_{t} \in \mathbb{R}^{(W_h \times W_w) \times T \times C_{e}}$ are respectively reshaped queries, keys and values.
According to the computational cost of the transformer models, we consider that DDCA should be embedded into the Clip-Recurrent Hub in a lightweight way to avoid affecting the propagation ability of fine-grained local features from the current clip.

Hence, we subsequently present the 2D-decoupled window-based attention with the strip pooling strategy to effectively achieve local-global interactions.
The horizontal keys $K_{h}$ can be naturally formulated as:
%
\begin{equation}
\label{equ:time_att_02}
    \begin{aligned}
    & K_{h}^{l}=[K_{h}^{1}, K_{h}^{2},..., K_{h}^{W}], \\
    & K_{h}^{g}={\rm Unfold}(\mathcal{P}_{h}(K_{h}^{l})), \\
    & K_{h} = [K_{h}^{l}, K_{h}^{g}], 
    \end{aligned}
\end{equation}
where $K_{h}^{l}$ denotes evenly partitioned non-overlapping horizontal strips, $\mathcal{P}_{h}$ denotes the horizontal strip pooling layer, and ${\rm Unfold}(\cdot)$ represents the unfold function along the strip.
The horizontal values $V_{h}$ can be similarly derived.
For horizontal attention, we have:
%
\begin{equation}
\label{equ:time_att_03}
    \begin{aligned}
    & Z_{h}^{i} = {\rm Attn}(Q_{h}^{i}, K_{h}^{i}, V_{h}^{i}), \\
    & Z_{h} = [Z_{h}^{1}, Z_{h}^{2},..., Z_{h}^{W}],
    \end{aligned}
\end{equation}
The vertically decoupled attention is similar. Finally, the output of these three parallel dimensions will be gathered: 
%
\begin{equation}
\label{equ:time_att_04}
    \begin{aligned}
    Z = \mathcal{P}_{t}(Z_t) + \mathcal{P}_{h,w}([Z_h, Z_v]),
    \end{aligned}
\end{equation}
where $\mathcal{P}_{t}$ and $\mathcal{P}_{h,w}$ are linear projections. 
Note that the above formulas omit the head dimension for simplicity. Therefore, our DDCA covers both global and local receptive fields by the cross-strip attention %
and the unfolded coarse-grained pooling strip in a single layer.
Different from the recent CSWin~\cite{dong2022cswin} which introduced cross-shaped window attention, our DDCA is put forward to process sequence data and bridge the connection between windows to enhance the local-global interaction via strip pooling mechanism. 
Our strategy also fundamentally differs from the axial attention~\cite{ho2019axial}, because it performs horizontal and vertical attention sequentially, while we compute the attention map in parallel in a decoupled 3D space. 

\noindent\textbf{Mix Fusion Feed Forward Network (MixF3N).}
The previous state-of-the-art VI-Trans~\cite{li2022towards,liu2021fuseformer} are equipped with overlapped-patch strategy (\ie soft split \& composite) to aggregate information from 
patches.
However, this is insufficient for the beyond-FoV estimation task, as it needs more realistic and fine-grained local details.
As previously shown in MiT~\cite{xie2021segformer}, convolutions could introduce local inductive bias in a direct way for the transformer backbone.
Thus, we introduce the two-branch depth-wise convolutions with kernels of different sizes in FFN to further enhance the free flow of sub-token information.
Suppose $A$ is the soft composite token vectors, the MixF3N is formulated as:
%
\begin{equation}
\label{equ:time_att_05}
    \begin{aligned}
    Z = {\rm MLP}({\rm GELU}({\rm SS}([\mathcal{C}_{3 \times 3}(A_{:\frac{C}{2}}),\mathcal{C}_{5 \times 5}(A_{\frac{C}{2}:})]))),
    \end{aligned}
\end{equation}
where $\mathcal{C}_{3 \times 3}$ and $\mathcal{C}_{5 \times 5}$ denote depth-wise convolutions, $A_{:\frac{C}{2}}$ and $A_{\frac{C}{2}:}$ are parallel features.
The MixF3N mixes a $3{\times}3$ and a $5{\times}5$ convolution into each FFN to model the spatial relationship among tokens in a multi-scale manner and reinforce the fine-grained local information extraction.
\begin{table*}[!t]
\renewcommand{\thetable}{I}
    \begin{center}
        \caption{\emph{Quantitative comparisons on KITTI360 online video inpainting for beyond-FoV scene estimation.} Inner FoV expansion results are not reported for SRN, as it primarily focuses on image outpainting. $E_{warp}^{*}$ denotes $E_{warp}$ scaled by a factor of $10^{-2}$. The best-performing results are indicated in \textbf{bold}, and the second-best results are \underline{underlined}.}
        \label{tab:fov_expansion}
        \input{tables/fov-expansion-v3}
    \end{center}
\end{table*}

\begin{figure}[!t]
  \centering
  \includegraphics[width=0.75\linewidth]{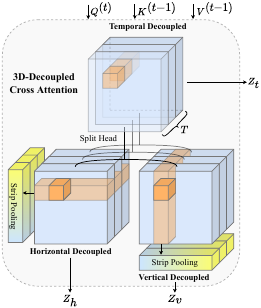}
  \caption{\emph{3D-Decoupled Cross Attention.} By decoupling the dimensions of time, width, and height, we are able to efficiently query the most correlated features from the past. With an additional non-local strip pooling window, the information flows flexibly in intersecting directions during spatial query.}
  \label{fig:cross_att}
\end{figure}

\subsection{Training Objective}
\label{sec:training_objective}
The FlowLens generator is end-to-end optimized with reconstruction loss, adversarial loss, and flow loss:
%
\begin{equation}
\label{equ:time_att_06}
    \begin{aligned}
    \mathcal{L} = \lambda_{rec} \cdot \mathcal{L}_{rec} + \lambda_{adv} \cdot \mathcal{L}_{adv} + \lambda_{flow} \cdot \mathcal{L}_{flow},
    \end{aligned}
\end{equation}
We empirically set the weights for different losses as: $\lambda_{rec}=0.01$, $\lambda_{adv}=0.01$ and $\lambda_{flow}=1$.
The reconstruction loss measures the $L1$ distance between the FoV-expanded sequence and the ground truth:
%
\begin{equation}
\label{equ:time_att_07}
    \begin{aligned}
    \mathcal{L}_{rec} = \Vert \mathbf{\hat{Y}}^{t} - \mathbf{Y}^{t} \Vert_{1}.
    \end{aligned}
\end{equation}
A discriminator~\cite{chang2019free} D is also equipped
to assist in the training of FlowLens for more realistic and structurally consistent FoV expansion results. The adversarial loss and the loss of discriminator D are formulated as:
%
\begin{equation}
\label{equ:time_att_08}
    \begin{aligned}
    \mathcal{L}_{adv} & = -E_{z \sim P_{\mathbf{\hat{Y}}^{t}}(z)}[D(z)], \\
    \mathcal{L}_{D} & = E_{x \sim P_{\mathbf{Y}^{t}}(x)}[{\rm ReLU}(1-D(x))] \\
    & + E_{z \sim P_{\mathbf{\hat{Y}}^{t}}(z)}[{\rm ReLU}(1+D(z))].
    \end{aligned}
\end{equation}
L1 loss is used for supervising the optical flow completion:
%
\begin{equation}
\label{equ:time_att_09}
    \begin{aligned}
    \mathcal{L}_{flow} = \Vert \hat{\mathbf{V}} - \mathbf{V}_{gt} \Vert_{1},
    \end{aligned}
\end{equation}
where $\mathbf{V}_{gt}$ is the ground-truth flow calculated from the video with a complete FoV. 

\begin{figure*}[!t]
  \centering
  \includegraphics[width=0.94\linewidth]{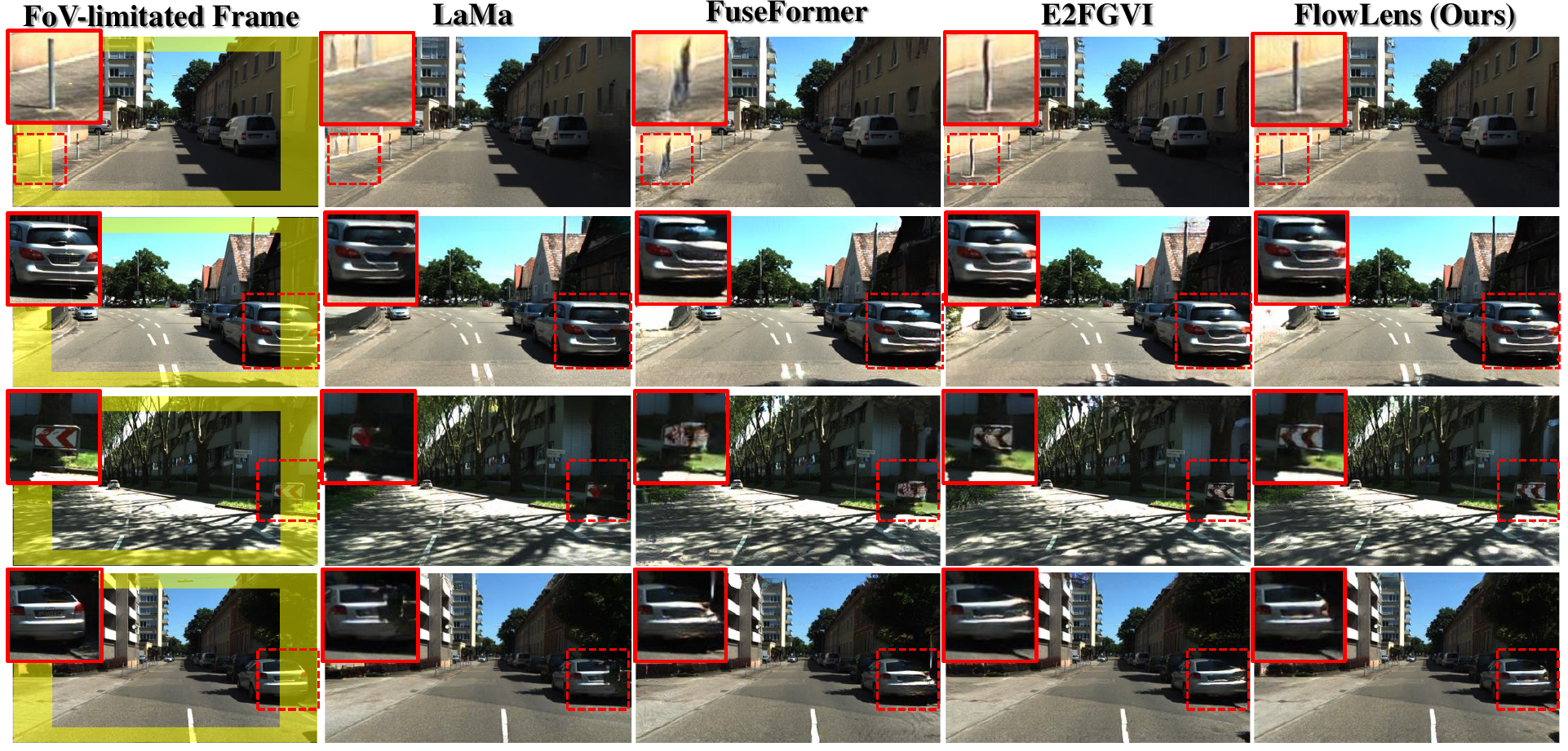}
  \caption{\emph{Qualitative comparison on KITTI360 outwards beyond-FoV scene estimation} with LaMa~\cite{suvorov2022resolution}, FuseFormer~\cite{liu2021fuseformer}, and E2FGVI~\cite{li2022towards}.}
  \label{fig:compare_kitti_out}
\end{figure*}

\section{Experiments}
\label{sec:experiments}

\subsection{Settings}
\label{sec:settings}
\noindent\textbf{Datasets.}
In this work, we experiment with three datasets, including two offline video inpainting datasets and one newly derived KITTI360 dataset for beyond-FoV scene estimation and perception.
\begin{compactitem}
\item[(1)] \textbf{YouTube-VOS}~\cite{xu2018youtube} contains of $3471$ videos for training, $474$ for validation, and $508$ for testing. We follow the original split for training and testing. 
\item[(2)] \textbf{DAVIS}~\cite{perazzi2016benchmark} provides $90$ videos for training and $60$ for testing. 
Following~\cite{li2022towards,liu2021fuseformer}, we use $50$ videos for testing the model that is trained on the YouTube-VOS.
\item[(3)] \textbf{KITTI360}~\cite{liao2022kitti} is used for beyond-FoV scene estimation and perception. Derived from the original KITTI360, it comprises a total of $76,000$ pinhole images as well as $76,000$ spherical images, which will be detailed in Sec.~\ref{sec:dataset}. We use ``seq10'' for testing. 
\end{compactitem}

\noindent\textbf{Metrics.}
For the online and offline video inpainting, we employ a comprehensive set of evaluation metrics to assess performance, including PSNR (Peak Signal-to-Noise Ratio), SSIM (Structural Similarity Index), VFID (Video-based Fr\'{e}chet Inception Distance), and $E_{warp}$ (Flow Warping Error). PSNR and SSIM are widely used for measuring the reconstructed image quality.
VFID~\cite{wang2018video} is used to compare the visual perceptual similarities between two input videos.
$E_{warp}$~\cite{lai2018learning} is used to measure the stability of the reconstructed video, where lower values indicate better temporal consistency. 
In the context of beyond-FoV perception, we use mIoU (mean Intersection over Union) to quantitatively assess the quality of semantic segmentation in both seen and unseen areas, offering a comprehensive evaluation of scene understanding. 
Training details can be found in the supplementary material.

\begin{table}[!t]
\renewcommand{\thetable}{II}
    \begin{center}
        %
        \caption{\emph{Quantitative comparisons on offline video inpainting}.} 
        \label{tab:video_inpainting}
        \input{tables/video-inpainting-v3}
    \end{center}
\end{table}

\subsection{Derive KITTI360 for Beyond-FoV Evaluation}
\label{sec:dataset}

Existing video editing datasets lack essential camera intrinsics for each video, and they do not include spherical images. To address this limitation, we leverage the KITTI360 dataset~\cite{liao2022kitti} to derive a dataset that facilitates training and evaluation for beyond-FoV scene reconstruction and perception.
Utilizing the calibrated camera intrinsics, we apply two camera models, the $f{-}tan\theta$ camera model and the $f{-}\theta$ camera model, to generate $5\%$, $10\%$, and $20\%$ Field of View (FoV) masks. These masks are constructed using the formulas ${\theta}_{x}{=}2arctan(\frac{x-{c}_{x}}{2{f}_{x}})$ and ${\theta}{=}{\frac{\sqrt{(x-{c}_{x})^2+(y-{c}_{y})^2}}{f}}$ for pinhole camera outward expansion and spherical camera inward expansion, respectively. This dataset enrichment is crucial for advancing beyond-FoV perception and scene understanding.
It is worth noting that during model evaluation on the KITTI360 dataset, we use only past frames, reflecting the real-world constraints of online applications for autonomous vehicles.

{\subsection{Architecture and Implementation Details}
\label{sec:implementation}
\noindent\textbf{Model Architecture.}
We implement the convolutional encoder and decoder architectures following~\cite{li2022towards,liu2021fuseformer}, where the channel dimension $C$ is set to $128$ for our standard model and $64$ for the small version. The optical flow completion network $\mathcal{F}$ is employed with MaskFlowNetS~\cite{zhao2020maskflownet}, and we initiate it with pre-trained weights on FlyingChairs~\cite{dosovitskiy2015flownet} and FlyingThings~\cite{mayer2016large} flow datasets to take advantage of the rich prior knowledge of optical flow. For FlowLens-s, we adopt pre-trained MaskFlowNetS to supervise SpyNet~\cite{ranjan2017optical} for speed consideration. The clip-recurrent transformer includes a mix focal transformer block embedded with a clip-recurrent hub and $8$ other transformer blocks for FlowLens and $4$ blocks for FlowLens-s. The head number $d$ of multi-head focal attention is set to $4$. The hidden 
dimension $C_{e}$ of the transformer is set to $512$, and $256$ for the small model. The input features of the transformer are split into $7{\times}7$ overlapping patches with $3{\times}3$ strides. The architecture of T-PatchGAN is the same as~\cite{li2022towards,liu2021fuseformer,zeng2020learning,chang2019free}.}

\noindent{\noindent\textbf{Training Details.}
For the offline video inpainting task, we set the batch size to $8$, the learning rate to $1{\times}10^{-4}$ to train $500k$ iterations following~\cite{li2022towards}, and the training image size is $432{\times}240$. The learning rate weight $\lambda_{rec}$ is set to $1$, and $\lambda_{adv}$ is set to $10^{-2}$. 
For online video inpainting of beyond-FoV scene reconstruction, we use a single NVIDIA RTX3090 with a batch size of $2$ and learning rate of $2.5 {\times} 10^{-5}$ to train all video inpainting models~\cite{li2022towards,liu2021fuseformer,zeng2020learning} and FlowLens for $500k$ iterations. The training pinhole image size is $432{\times}240$ and the size of the spherical image is $336{\times}336$. During training, the number of local frames $T_{lf}$ is $5$, and the number of past reference frames $T_{pf}$ is $3$. For FlowLens, we use serialized loading of training logic and empty the clip buffer when switching to a new video to avoid interference from irrelevant clips. 
Ablations are performed on the inward expansion with a batch size of $2$ for $250k$ iterations.}

\begin{figure*}[!t]
  \centering
  \includegraphics[width=0.90\linewidth]{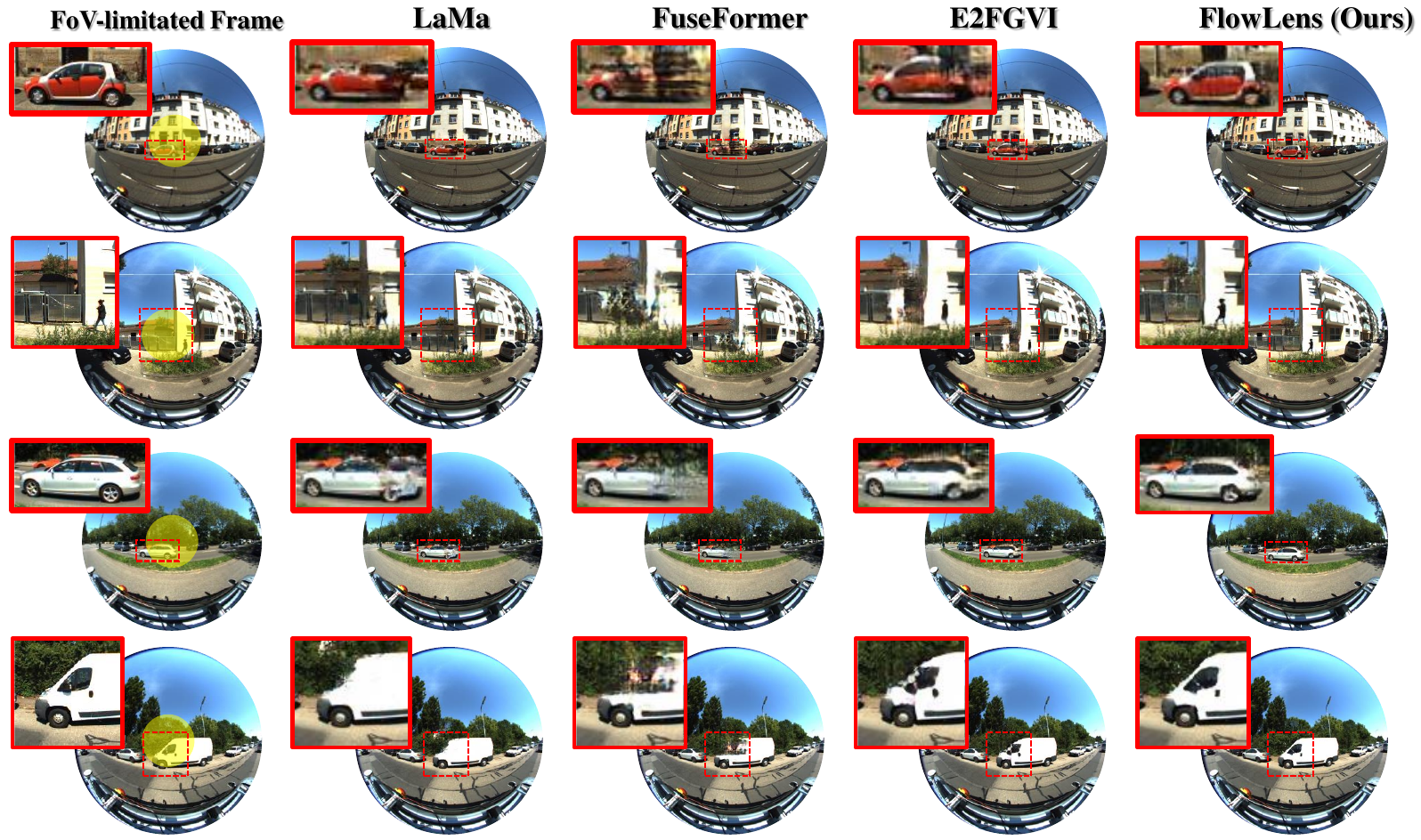}
  \caption{\emph{Qualitative comparison on KITTI360 inwards beyond-FoV scene estimation} with LaMa~\cite{suvorov2022resolution}, FuseFormer~\cite{liu2021fuseformer}, and E2FGVI~\cite{li2022towards}.}
  \label{fig:compare_kitti_in}
  \vspace{-1em}
\end{figure*}

\begin{figure*}[!t]
  \centering
  \includegraphics[width=1.0\linewidth]{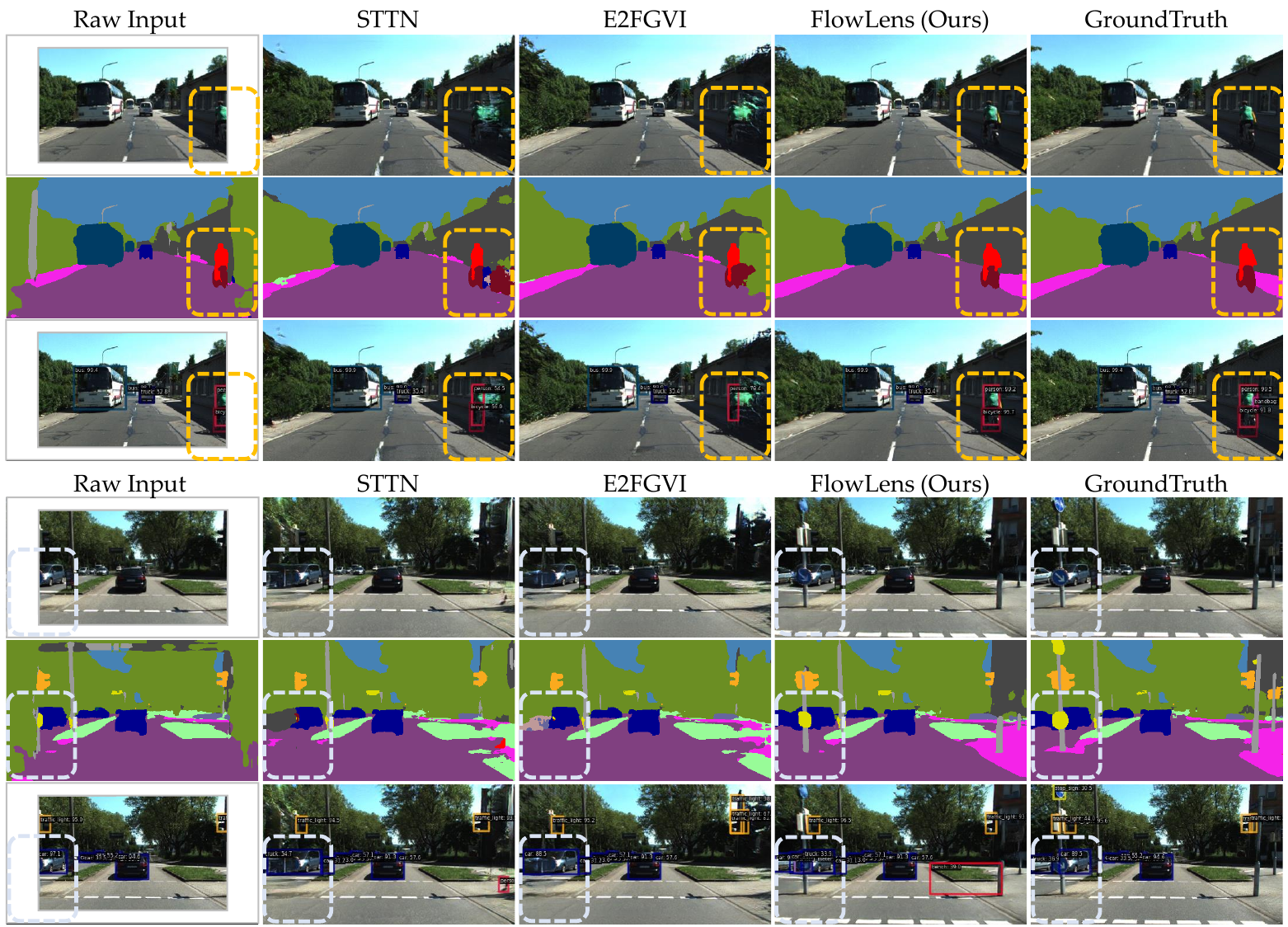}
  \caption{\emph{Qualitative results of semantic segmentation and object detection beyond the field of view, utilizing a pre-trained segmentation method (SegFormer~\cite{xie2021segformer}) and a detector (Faster R-CNN~\cite{ren2015faster}), applied to RGB images obtained from the test set of different online video inpainting pipelines captured using pinhole cameras.}}
  \label{fig:compare_kitti_out_percep}
  \vspace{-1em}
\end{figure*}

\begin{figure*}[!t]
  \centering
  \includegraphics[width=0.85\linewidth]{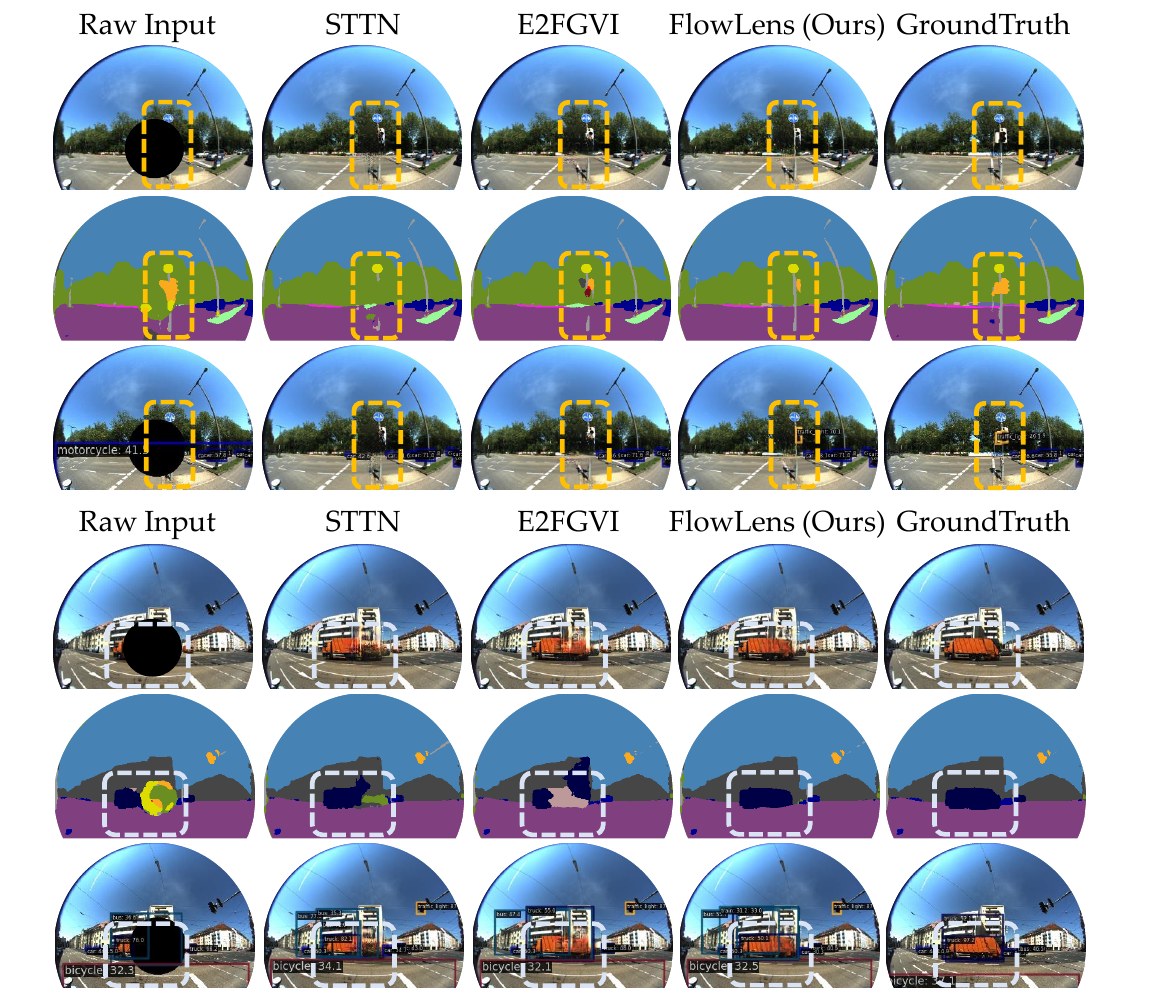}
  \caption{\emph{Qualitative results of semantic segmentation and object detection beyond the field of view, utilizing a pre-trained segmentation method (SegFormer~\cite{xie2021segformer}) and a detector (Faster R-CNN~\cite{ren2015faster}), applied to RGB images obtained from the test set of different online video inpainting pipelines captured using spherical cameras.} Please note that the lower half of the image has been cropped from the results as it mainly consists of the camera mount device.}
  \label{fig:compare_kitti_in_percep}
  \vspace{-1em}
\end{figure*}

\subsection{Quantitative Comparisons}
\label{sec:comparison}

We report quantitative comparison results on online and offline video inpainting, as well as beyond-FoV scene understanding. With ``-s'' indicates the small version of our method, and ``+'' denotes that horizontal and vertical flip augmentation are equipped during inference.

\subsubsection{Online Video Inpainting for Beyond-FoV Scene Reconstruction}
In the context of online video inpainting for beyond-FoV scene reconstruction, all methods considered allow only past video frames to serve as references. Our evaluation encompasses recent image inpainting techniques~\cite{suvorov2022resolution,liu2022reduce}, the image outpainting method SRN~\cite{wang2019wide}, and state-of-the-art video inpainting models~\cite{li2022towards,liu2021fuseformer,zeng2020learning}. To ensure a fair comparison, all video inpainting models and FlowLens adhere to the same training settings.
As illustrated in Tab.~\ref{tab:fov_expansion}, FlowLens demonstrates its superior performance. When compared to the best-performing video inpainting model E2FGVI~\cite{li2022towards}, FlowLens surpasses it by substantial margins, achieving a $0.68dB$ higher PSNR for outer FoV expansion and a $0.78dB$ higher PSNR for inner FoV expansion. Furthermore, when compared to the top image inpainting model, LaMa~\cite{suvorov2022resolution}, our method also exhibits significant improvements.
FlowLens clearly outperforms prior approaches in beyond-FoV scene reconstruction using the online video inpainting paradigm, establishing a promising baseline for this innovative approach. Notably, the smaller version of FlowLens outperforms all other VI-Trans models while consuming only $37\%$ of the FLOPs compared to E2FGVI~\cite{li2022towards}. It achieves exceptional results with a PSNR of $19.68dB$ and $36.17dB$ for the outer and inner sets, respectively. 
Additionally, its inference takes an average runtime of $0.023$ seconds per frame, making it ideally suited for real-time autonomous applications.

\begin{table}[t]
    \begin{center}
        \vskip 2ex
        \caption{{\emph{Half-precision speed performance on edge computing platform.}}}
        \label{tab:revise_jetson}
        \input{tables/revise_jetson_R2}
    \end{center}
\end{table}

To further illustrate the deployability of FlowLens on intelligent vehicles, we conduct experiments on NVIDIA's edge computing platform Jetson Orin NX 16GB. During these testing, we adjust the model to half-precision calculation, and the input local frame window is set to $5$. We extend the pinhole image outward by $10\%$ of the field of view. In addition, we compare the half-precision calculation speed and accuracy of E2FGVI~\cite{li2022towards} under the same onboard setting. {As shown in Tab.~\ref{tab:revise_jetson}, the FlowLens-S model achieves a running speed of $27.31$ FPS ($0.037$ s/frame) with higher accuracy ($19.04dB$  \textit{vs.} $18.74dB$ ), proving that its computational overhead is acceptable for real-world vehicle applications.}   

\subsubsection{Offline Video Inpainting for Video Editing}
For the offline video inpainting task, we conduct comparisons with previously top-performing video inpainting models~\cite{li2022towards,liu2021fuseformer,kim2019deep,xu2019deep,zeng2020learning,gao2020flow,lee2019copy,chang2019learnable} under the same test conditions and mask configurations, following the methodology outlined in~\cite{li2022towards,liu2021fuseformer}.
As demonstrated in Tab.~\ref{tab:video_inpainting}, FlowLens exhibits state-of-the-art performance, particularly excelling in terms of the quality and structural similarity of the reconstructed videos. On the YouTube-VOS dataset, our method achieves remarkable results with a PSNR of $34.23dB$ and an SSIM of $0.9731$. Additionally, on the offline video inpainting dataset DAVIS, FlowLens+ surpasses the previous best-performing method, achieving a PSNR of $33.47dB$ and an SSIM of $0.9750$. 
The outstanding results of our method in both online and offline video inpainting tasks underscore the superiority of the proposed clip-recurrent transformer, emphasizing its potential to enhance the perception range of various autonomous and intelligent vehicle systems.

\begin{table}[!t]
    \begin{center}
        \vspace{1.0em}
        \caption{\emph{Beyond field-of-view semantics for Pinhole camera.}}
        \label{tab:beyond-semantics-pinhole}
        \input{tables/beyond-semantics-pinhole}
    \end{center}
\end{table}

\subsubsection{Online Video Inpainting for Beyond-FoV Scene Understanding}
We demonstrate that an individual FlowLens model can effectively enhance high-level vision tasks beyond the Field of View (FoV) for autonomous applications, such as semantic segmentation, without the need for joint training or additional labels.

To evaluate the improvement in perceptual ability due to visual completion, it is necessary to determine the upper limit of perceptual ability within the complete field of view. We acquire pseudo-labels for perceptual results by utilizing full FoV observations. The better the visual completion effect, the closer the perceived result aligns with these predefined pseudo-labels. Specifically, we employ the KITTI360 dataset and utilize a pre-trained semantic segmentation model (SegFormer~\cite{xie2021segformer}) to generate pseudo-labels from the original complete vision captures. 
Notably, we directly evaluate images processed by FlowLens and other video inpainting models under online training and inference logic, without pre-training on the segmentation task. The beyond-FoV semantic results for pinhole cameras using various video inpainting methods are presented in Tab.~\ref{tab:beyond-semantics-pinhole}. Our method demonstrates a significant performance improvement compared to others, with an increase of $18.99\%$ in mIoU for unseen regions, and even a $2.30\%$ increase in mIoU for previously seen regions. Similar trends are observed in tests involving spherical cameras, as shown in Table~\ref{tab:beyond-semantics-sphere}, where FlowLens achieves a performance gain of $13.63\%$ in mIoU for unseen regions and a $4.10\%$ increase in mIoU for previously seen regions. We argue that this indicates that FlowLens can effectively complete unseen regions with coherent context, thereby benefiting the semantic segmentation of the original regions. 

\begin{table}[t]
    \begin{center}
        \caption{\emph{Comparison on beyond field-of-view detection.}}
        \label{tab:beyond_detection}
        \input{tables/beyond-detection-pinhole}
    \end{center}
\end{table}

~~\textit{4) Online Video Inpainting for Beyond-FoV Object Detection:} 
To evaluate the efficacy of object detection beyond the field of view, we engage in detailed quantitative experiments. The core of our experimental setup involves the Faster-RCNN~\cite{ren2015faster} model, which is trained using the COCO~\cite{lin2014microsoft} dataset. Our first step is to determine the ground truth across the entire field of view, enabling us to create pseudo-labels for object detection. Following this, we conduct inference tests on images encompassing both the limited field of view and an expanded field of view, extended by $10$\% through various online video inpainting methods~\cite{zeng2020learning, li2022towards} and the proposed FlowLens.

Our evaluation focuses on two key metrics: Average Recall (AR$^{\text{box}}$) and Average Precision (AP$^{\text{box}}$) for bounding box Intersection over Union (IoU) values ranging from $0.50$ to $0.95$, at intervals of $0.05$, with a maximum of $100$ detections. These metrics are chosen to provide a balanced view of the model's performance, reflecting its precision across a spectrum of positional accuracy criteria. Additionally, we calculate the average accuracies, AP$^{\text{box}}_{50}$ and AP$^{\text{box}}_{75}$ at specific IoU thresholds $0.50$ and $0.75$, respectively.

As shown in Tab.~\ref{tab:beyond_detection}, the results from these experiments are revealing. When compared to frames limited by the field of view, all tested online video inpainting methods enhance the accuracy of object detection beyond the field of view. Notably, FlowLens demonstrates remarkable improvements in both AR$^{\text{box}}$ and AP$^{\text{box}}$, by $7.4$\% and $6.5$\% respectively. This significant advancement underscores the effectiveness of FlowLens in enhancing object detection capabilities outside the field of view in intelligent vehicle systems.

\begin{table}[!t]
    \begin{center}
        \caption{\emph{Beyond field-of-view semantics for Spherical camera.}}
        \label{tab:beyond-semantics-sphere}
        \input{tables/beyond-semantics-sphere}
    \end{center}
\end{table}

\subsection{Qualitative Results}
\subsubsection{User Study}
\begin{figure}[!t]
    \centering
    \includegraphics[width=1\columnwidth]{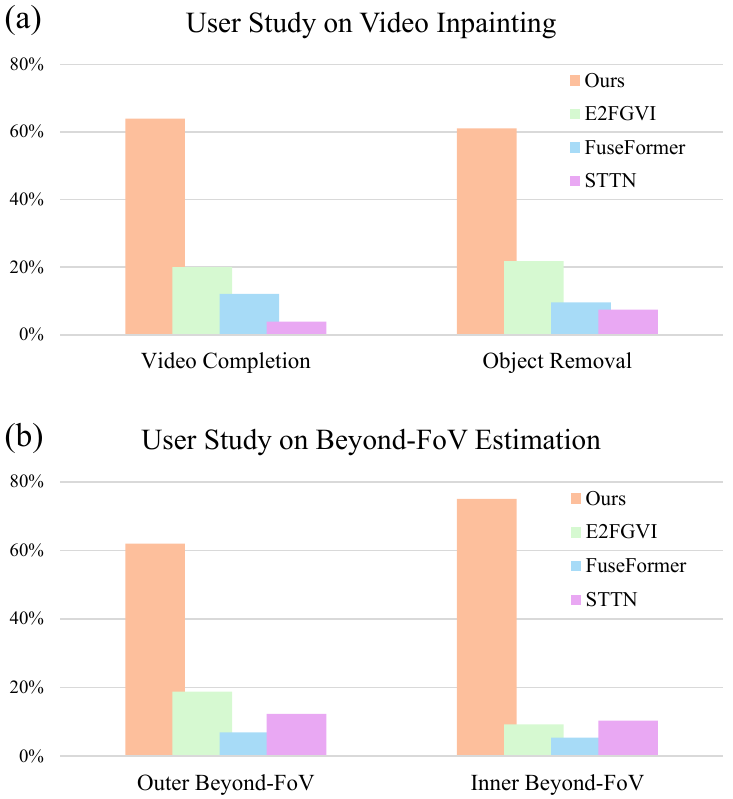}
    \caption{User study results. The vertical axis indicates the percentage of ranking first among ${28}$ viewers of ${40}$ videos on (a) offline video inpainting for video editing and (b) online video inpainting for beyond-FoV scene reconstruction.}
    \label{fig:sup_user_study}
\end{figure}

For a comprehensive comparison, a user study was conducted with top-performing methods~\cite{liu2021fuseformer,li2022towards,zeng2020learning}.
To be specific, we randomly sample $20$ videos from DAVIS~\cite{perazzi2016benchmark} to evaluate offline video inpainting ($10$ for video completion, and $10$ for object removal). $20$ videos are also randomly sampled from KITTI360 for evaluating online video inpainting for beyond-FoV scene reconstruction ($10$ for outward estimation, and $10$ for inner estimation). $28$ volunteers are invited to participate in the survey. Each volunteer simultaneously watches the outputs of the $4$ methods, as well as the original video input. Volunteers need to conduct a total of $40$ trials. 
For a fair comparison, there is no time limit for trials, and each video can be paused and replayed at any time. The statistics are shown in Fig.~\ref{fig:sup_user_study}, our method clearly performs better for these two types of tasks, demonstrating that the proposed method can produce more visually pleasing results. Interestingly, we find that STTN outperforms FuseFormer on the beyond-FoV setting, suggesting an essential difference in the model requirements of the two tasks. We consider that the Clip-Recurrent Transformer of FlowLens is able to mine potential visual cues in past iterations, thus gaining advantages on both tasks at the same time.

\subsubsection{Qualitative Comparisons}
\begin{figure}[!t]
  \centering
  \includegraphics[width=1.0\linewidth]{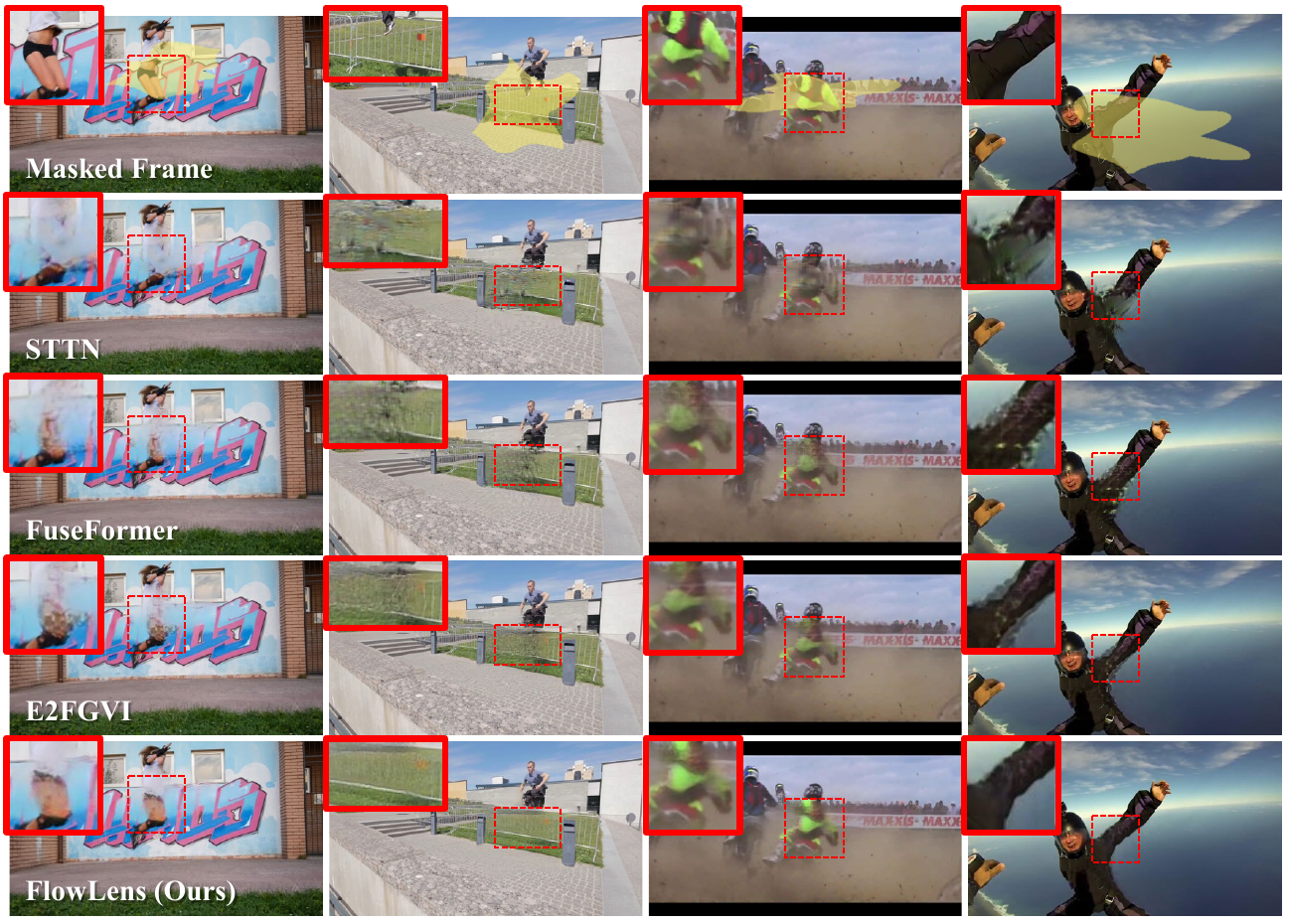}
  \caption{\emph{Qualitative comparison for offline video inpainting} against recent VI-Trans~\cite{li2022towards,liu2021fuseformer,zeng2020learning} on DAVIS~\cite{perazzi2016benchmark} and YouTube-VOS~\cite{xu2018youtube}.}
  \label{fig:compare_vi}
\end{figure}

We conduct qualitative comparisons with the competitive LaMa~\cite{suvorov2022resolution} and advance video inpainting models~\cite{li2022towards,liu2021fuseformer,zeng2020learning}.  
Fig.~\ref{fig:compare_kitti_out} and Fig.~\ref{fig:compare_kitti_in} show the results of outwards and inwards beyond-FoV scene estimation, respectively. Fig.~\ref{fig:compare_kitti_out_percep} and Fig.~\ref{fig:compare_kitti_in_percep} provide examples to compare our method with previous approaches in terms of high-level perception performance beyond the field of view, including semantic segmentation and object detection. In Fig.~\ref{fig:compare_vi} we present the qualitative comparisons on the offline video inpainting task. FlowLens takes advantage of the proposed unique clip-recurrent mechanism to efficiently utilize the redundant textures and structures in the same input conditions, thus also achieving competitive performance in video completion tasks. Observations can be made as follows:
(1) In Fig.~\ref{fig:compare_kitti_out} and Fig.~\ref{fig:compare_kitti_in}, the results demonstrate that the popular image inpainting method LaMa~\cite{suvorov2022resolution} fails to effectively address the issue of estimating the presence of objects outside of the view frustum. When the objects outside the FoV are completely out of focus, it is not possible for image inpainting methods to infer the presence of the semantic information from the current frame. For example, in Fig.~\ref{fig:compare_kitti_out}, LaMa neglects the guardrails outside of the frame, and in Fig.~\ref{fig:compare_kitti_in}, it fails to redraw the car window. 
(2) In comparison to advanced VI-Trans~\cite{li2022towards,liu2021fuseformer,zeng2020learning} in Fig.~\ref{fig:compare_kitti_out} and Fig.~\ref{fig:compare_kitti_in}, FlowLens is able to propagate more faithful textures and structures to the filling area, achieving competitive qualitative performance, especially in areas with rich semantic content, such as vehicles, pedestrians, and traffic signs. This is crucial for maintaining perception and safety in areas outside of the visual field.
(3) In Fig.~\ref{fig:compare_kitti_out_percep} and Fig.~\ref{fig:compare_kitti_in_percep}, as demonstrated in the fourth column, our method consistently exhibits superior performance compared to existing video inpainting techniques in complex driving scenarios. The enhancements achieved through the clip-recurrent transformer, especially in global information aggregation and hybrid propagation, result in improved reconstruction quality and greater accuracy in predicting high-level tasks. In Fig.~\ref{fig:compare_kitti_out_percep}, FlowLens effectively reconstructs the cyclist and the blue arrow sign located outside the field of view. It accurately segments their semantics and performs object detection. In Fig.~\ref{fig:compare_kitti_in_percep}, FlowLens successfully reconstructs the road poles and trucks in the blind spot, with precise segmentation and detection. This effectively expands the visual perception boundaries of the vehicle, thereby enhancing the intelligent vehicles' cognitive capabilities to the surrounding environment and driving safety.

In summary, FlowLens outperforms advanced VI-Trans on both inwards and outwards beyond-FoV scene estimation and perception using the online video inpainting fashion, while also presenting a competitive offline video inpainting advantage, demonstrating the effectiveness of our approach. 

\begin{table}[!t]
    \begin{center}
        \caption{\emph{Ablation studies on clip-recurrent hub.}}
        \label{tab:ablation-recurrent}
        \input{tables/ablation-recurrent-v2}
    \end{center}
\end{table}

\begin{table}[!t]
    \begin{center}
        \caption{\emph{Ablation studies on various cross-attention mechanism.}}
        \label{tab:ablation-cross}
        \input{tables/ablation-cross-v2}
    \end{center}
\end{table}

\subsection{Ablations}
\label{sec:ablations}
We conduct ablations on the clip-recurrent hub, cross attention, MixF3N, and flow completion, on the KITTI360-EX Spherical track with $250k$ training iterations, and the results averaged across metrics are reported.

\begin{table}[h]
    \begin{center}
        \caption{\emph{Effectiveness of FlowLens framework with ProPainter baseline~\cite{zhou2023propainter}.}}
        \label{tab:propainter}
        \input{tables/revise_propainter}
    \end{center}
    \vskip -3ex
\end{table}

\noindent\textbf{Study of the effectiveness of FlowLens framework.} 
To further reveal the effectiveness of the FlowLens framework, we integrate the latest state-of-the-art baseline, ProPainter~\cite{zhou2023propainter}, into our study.
Our approach involves conducting comparative experiments on the KITTI-360 pinhole camera, where we expand the field of view by $5$\%, $10$\%, and $20$\%, respectively. We align our methodology with ProPainter's training protocol, initially using RAFT~\cite{teed2020raft} to compute optical flow pseudo-labels, followed by training with $5$ local frames and $3$ reference frames. The enhancements introduced in the FlowLens (ProPainter) model are pivotal, comprising 1) the Clip-Recurrent Hub, designed to aggregate features across a broader time window, and 2) the Mix-F3N, aimed at enhancing fine-grained local spatial cues.
The results, as depicted in Tab.~\ref{tab:propainter}, indicate that the FlowLens (ProPainter) model incurs only a minor increase in computational time. Significantly, it exhibits superior accuracy performance in comparison to the baseline at all tested field of view expansion rates, achieving an average improvement of $0.41$dB PSNR. These results further validate the efficacy of the FlowLens framework.

\noindent\textbf{Study of the clip-recurrent hub.}
We explore whether all layers are required to equip the clip recurrent hubs, and if not, what positions are most efficient. Tab.~\ref{tab:ablation-recurrent} shows that regardless of where the hub is added, it performs better than the setting without the hub. Interestingly, all layers with cross attention are not necessary, and introducing the hub at the early stage of the transformer works better. We consider that the early fusion can enable the subsequent mix focal transformer to better handle relevant visual cues without causing confusion for the feature extraction of the current clip. Besides, the proposed clip-recurrent hub only requires negligible cost at $1.4\%$ FLOPs and $2.9\%$ parameters of the entire model.

\noindent\textbf{Study of cross attention mechanism.}
Tab.~\ref{tab:ablation-cross} compares the effects of different cross-attention mechanisms on beyond-FoV estimation. We observe that the proposed DDCA works satisfactorily with the hub, especially considering its moderate computational complexity. 
What is the secret of the DDCA module? 
To answer this question, we further conduct ablations on decoupling and strip pooling as shown in Tab.~\ref{tab:reb_strip_pooling}. The temporal-decoupled attention is able to search and propagate relevant features from a smaller spatial region along the time axis, while spatial-decoupled strip attention is able to search the entire spatial region in orthogonal local strip windows. Further, via orthogonal strip pooling, parallel local windows are bridged, which strengthens the global context. The results demonstrate that when these two features are both in place, the DDCA module achieves satisfactory performance.

\begin{table}[t]
    \begin{center}
        \caption{\emph{Dive into DDCA module.} Decoupling and Strip Pooling indicate 3D attention decoupling and orthogonal strip pooling in DDCA, respectively.}
        \label{tab:reb_strip_pooling}
        \input{tables/ablations-ddca}
        \vspace{1em}
    \end{center}
    \vskip -3ex
\end{table}

\begin{figure}[!t]
  \centering
  \includegraphics[width=1.0\linewidth]{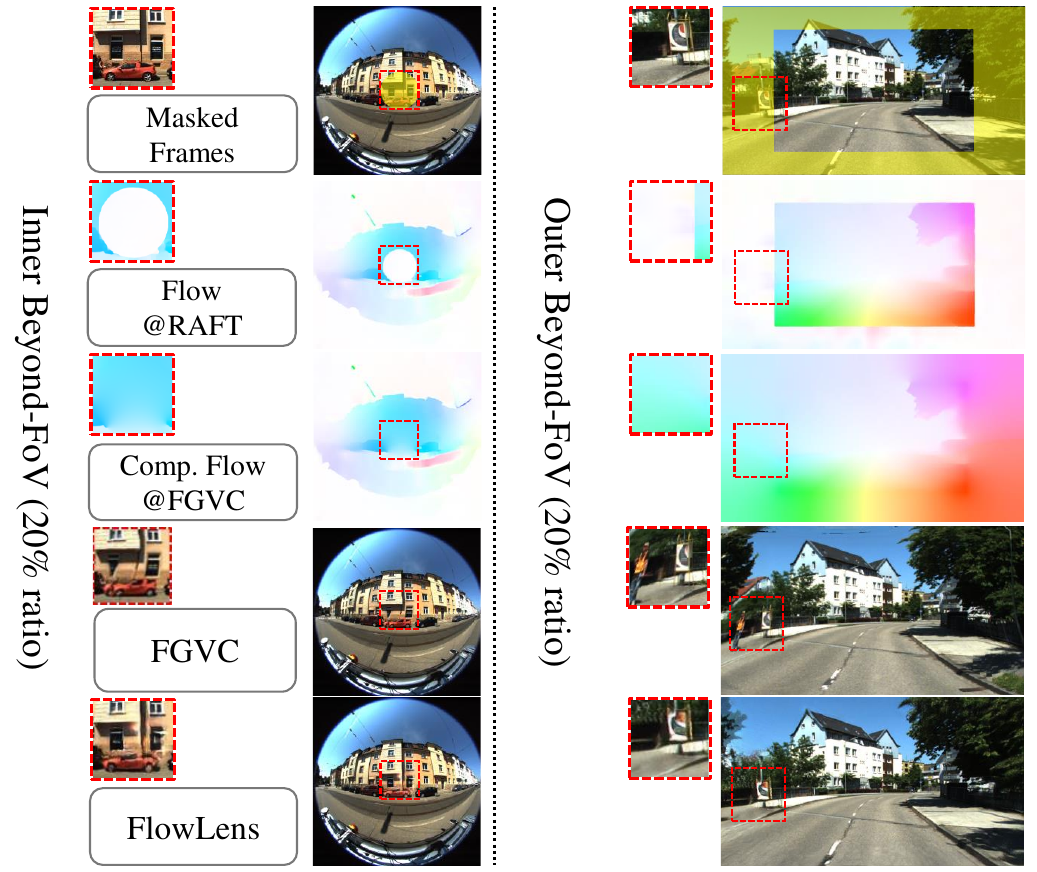}
  \caption{Comparing method~\cite{gao2020flow} that only rely on explicit optical flow propagation with FlowLens' hybrid propagation.}
  \label{fig:compare_fgvc}
\end{figure}

\begin{table}[!t]
    \begin{center}
        \caption{\emph{Ablation studies on flow-guided feature propagation.}}
        \label{tab:ablation-flow}
        \input{tables/ablation-flow}
    \end{center}
    \vskip -3ex
\end{table}

\noindent\textbf{Study of flow-guided feature propagation.}
Tab.~\ref{tab:ablation-flow} shows that performance drops dramatically when optical flow guidance or DCN compensation is removed, presumably because the accurate motion field and sampling error compensation are both important for utilizing multi-frame information, especially for beyond-FoV estimation. The result confirms that we further improve the performance by incorporating the strong flow completion network~\cite{zhao2020maskflownet} instead of the previous SpyNet~\cite{ranjan2017optical}.
We also explore the effect of optical flow resolution $R$ on local pixel propagation. Tab.~\ref{tab:reb_flow_res} shows the results of the flow resolution study. When $R$ is higher, the flow density is higher, but the absolute value of motion is larger and difficult to complete, which leads to a performance drop as well as time consumption, thus the choice needs to be weighed. 
Our model achieves satisfactory performance at 
$R{=}\frac{1}{4}$,
consistent with our setting.

\begin{table}[t]
    \begin{center}
        \caption{\emph{The flow resolution study on inner FoV expansion set.}}
        \label{tab:reb_flow_res}
        \input{tables/ablation-flow-res}
        \vspace{-1.0em}
    \end{center}
\end{table}

In order to closely reveal the advantages of the feature propagation approach proposed by FlowLens, We present a qualitative comparison with the explicit flow propagation method~\cite{gao2020flow} on the beyond-FoV task. 
As shown in Fig.~\ref{fig:compare_fgvc}, FGVC performs well on the inward FoV expansion with reasonable predictions but is challenged in extending the FoV to the outside. This is due to FGVC leveraging explicit flow propagation for completion, which can improve generalization to different data but is hampered by inaccurate flow completion in terms of outward expansion with fewer constraints. In contrast, FlowLens combines explicit flow-guided propagation with implicit transformer-based feature propagation, therefore it can efficiently transmit relevant temporal and spatial cues, achieving high-fidelity beyond-FoV estimation.

\noindent\textbf{The effectiveness of MixF3N in FlowLens.}
In Tab.~\ref{tab:ablation-mix}, we ablate the variants of FFN. Previous VI-Trans work~\cite{li2022towards,liu2021fuseformer} propose to use F3N instead of FFN, however, we find its minor improvement for beyond-FoV estimation.
Our results also show that Mix-FFN~\cite{xie2021segformer} does improve the transformer's ability of implicit feature propagation.
The newly introduced MixF3N, which combines the patch overlapping strategy and the dual-branch mix convolution together, further boosts the state-of-the-art score.

\begin{table}[t]
    \begin{center}
        \caption{\emph{Ablation studies on Mix Fusion Feed Forward Network.}}
        \label{tab:ablation-mix}
        \input{tables/ablation-mix}
    \end{center}
    \vskip -3ex
\end{table}

\subsection{Limitations}
The results of outward expansion for pinhole cameras are not as satisfactory as the inward expansion for spherical cameras at the same FoV expansion rate. This suggests that scene estimation and perception beyond the FoV in an outward direction remains a challenging task. This is due to the unidirectional constraint, the larger practical area to be filled brought by the $f{-}\theta$ camera model, and the larger displacement motion field compared to the inward expansion situation. 

In the future, we intend to explore beyond-FoV vision and perception in the context of multi-modal fusion. 
Additionally, we plan to investigate the feasibility and practical value of beyond-FoV sensing in 3D vision, such as using LiDAR point clouds and event cameras. This approach aims to make algorithms and information-collecting sensors more complementary to each other, thereby addressing these limitations.

\section{Conclusion}
\label{sec:clu}
In this study, we delve into the concept of expanding vision and perception capabilities beyond the limitations of a vehicle-mounted camera's field-of-view (FoV). Our aim is to leverage past spatiotemporal information to extend the vision and perception of autonomous agents beyond the physical boundaries of the camera's FoV. To tackle this challenge and bridge the gap between limited vision capture and beyond-FoV high-level recognition, we introduce FlowLens, an innovative clip-recurrent transformer architecture. FlowLens excels at efficiently querying past 3D coherent visual cues, facilitating a deeper understanding of scenes that are outside the immediate view of the camera.
Extensive experiments and user studies demonstrate the remarkable capabilities of FlowLens. It outperforms existing methods, achieving state-of-the-art quantitative and qualitative results across various tasks, including offline/online video inpainting, and the emerging field of beyond-FoV scene parsing. These results highlight FlowLens' potential to significantly enhance perception and scene understanding in the context of autonomous vehicles and intelligent systems.

We believe that FlowLens represents a crucial step toward breaking the physical limitations of information-collecting sensors. By expanding our perception capabilities beyond the camera's FoV, we can improve the safety, efficiency, and reliability of autonomous vehicles and other intelligent applications. We anticipate that FlowLens will stimulate further research and innovation in this exciting area.

{\small
\bibliographystyle{IEEEtran}
\bibliography{egbib}
}

\input{supp}

\end{document}

%% file: codes/recurrent-code.tex
\algsetup{linenosize=\tiny} \scriptsize %
    \PyComment{\# qkv\_proj: linear project layer} \\
    \PyComment{\# linear\_k, linear\_v: update modules} \\
    \PyComment{\# fusion: fusion project layer} \\
    \PyCode{class ClipRecurrentHub():} \\
        \PyComment{\quad \# clip buffer} \\
        \PyCode{\quad self.c\_k = None} \\
        \PyCode{\quad self.c\_v = None} \\

        \PyComment{\quad \# cross attention} \\
        \PyCode{\quad self.att = Decoupled3DAttention()} \\ 
        
        \PyCode{\quad def forward(x):} \\
        \PyCode{\qquad q,k,v = self.qkv\_proj(x)} \\ 
        \PyComment{\qquad \# update last clip cache} \\
        \PyCode{\qquad c\_k = self.linear\_k(self.c\_k)} \\ 
        \PyCode{\qquad c\_v = self.linear\_v(self.c\_v)} \\ 

        \PyComment{\qquad \# 3D-decoupled cross attention} \\
        \PyCode{\qquad c\_x = self.att(q, c\_k, c\_v)} \\
        \PyCode{\qquad c\_x = self.att\_proj(c\_x)} \\
        
        \PyComment{\qquad \# fusion} \\
        \PyCode{\qquad res\_x = self.fusion(cat(x, c\_x))} \\
        \PyCode{\qquad z = x + res\_x} \\
        
        \PyComment{\qquad \# cache new clip values} \\
        \PyCode{\qquad self.c\_k = k.detach()} \\
        \PyCode{\qquad self.c\_v = v.detach()} \\
        \PyCode{\qquad return z} \\

%% file: tables/fov-expansion-v3.tex
\resizebox{\textwidth}{!}{ 
\setlength{\tabcolsep}{1mm}{ %
\begin{tabular}{c|l|cccc:cccc|c|c}

\hline
\multicolumn{2}{c|}{\multirow{3}{*}{\textbf{Method}}} & \multicolumn{8}{c|}{\textbf{KITTI360}} & \multicolumn{2}{c}{\textbf{Efficiency}} \\
\cline{3-12}
\multicolumn{1}{c}{}&& \multicolumn{4}{c:}{Outer FoV Expansion} & \multicolumn{4}{c|}{Inner FoV Expansion} & \multirow{2}{*}{FLOPs} & \multirow{2}{*}{\makecell[c]{Runtime \\ (s/frame)}}\\ %

\cline{3-10}
\multicolumn{1}{c}{}&& PSNR$\uparrow$ & SSIM$\uparrow$ & VFID$\downarrow$ & $E_{warp}^{*}\downarrow$ & PSNR$\uparrow$ & SSIM$\uparrow$ & VFID$\downarrow$ & $E_{warp}^{*}\downarrow$ & &\\

\hline
\hline
Image Inpainting & LaMa~\cite{suvorov2022resolution} & 18.98 & 0.9073 & 0.355 & 0.5802 & 31.69 & 0.9876 & 0.026 & 0.3729 & 778.4G & 0.088  \\

\& & PUT~\cite{liu2022reduce} & 18.70 & 0.9020 & 0.321 & 0.7782 & 26.26 & 0.9657 & 0.101 & 0.5326 & - & 14.83  \\

Outpainting & SRN~\cite{wang2019wide} & 16.10 & 0.8439 & 0.479 & 0.6065 & - & - & - & - & - & 0.032 \\

\hline

\multirow{3}{*}{Video Inpainting} & STTN~\cite{zeng2020learning} & 18.73 & 0.9078 & 0.371 & 0.6537 & 35.84 & 0.9909 & \underline{0.023} & 0.3729 & 885.6G & \textbf{0.019} \\

& FuseFormer~\cite{liu2021fuseformer} & 18.91 & 0.9121 & 0.380 & 0.7066 & 34.78 & 0.9880 & 0.034 & 0.3841 & \underline{446.9G} & 0.033 \\

& E2FGVI~\cite{li2022towards} & 19.45 & 0.9229 & 0.313 & 0.5816 & 35.91 & 0.9904 & 0.029 & 0.3728 & 560.0G & 0.038 \\

\hline

\multirow{3}{*}{Ours} & FlowLens & \underline{20.13} & \underline{0.9314} & \textbf{0.281} & 0.5347 & \underline{36.69} & \underline{0.9916} & 0.027 & 0.3672 & 586.6G & 0.049 \\

& FlowLens-s & 19.68 & 0.9247 & \underline{0.300} & \underline{0.5084} & 36.17 & \underline{0.9916} & 0.030 & \underline{0.3660} & \textbf{207.2G} & \underline{0.023} \\

\rowcolor{gray!20}
& FlowLens+ & \textbf{20.50} & \textbf{0.9322} & \underline{0.300} & \textbf{0.3913} & \textbf{37.38} & \textbf{0.9926} & \textbf{0.019} & \textbf{0.3589} & 586.6G & 0.148 \\

\hline

\end{tabular}
}
}

%% file: tables/video-inpainting-v3.tex
\resizebox{\columnwidth}{!}{
\setlength{\tabcolsep}{0.25mm}{    %
\begin{tabular}{l|cccc:cccc}

\hline
& \multicolumn{4}{c:}{\textbf{YouTube-VOS}} & \multicolumn{4}{c}{\textbf{DAVIS}} \\

\cline{2-9}
\textbf{Method} & PSNR$\uparrow$ & SSIM$\uparrow$ & VFID$\downarrow$ & $E_{warp}^{*}\downarrow$ & PSNR$\uparrow$ & SSIM$\uparrow$ & VFID$\downarrow$ & $E_{warp}^{*}\downarrow$ \\

\hline
\hline
VINet~\cite{kim2019deep} & 29.20 & 0.9434 & 0.072 & 0.1490 & 28.96 & 0.9411 & 0.199 & 0.1785 \\

DFVI~\cite{xu2019deep} & 29.16 & 0.9429 & 0.066 & 0.1509 & 28.81 & 0.9404 & 0.187 & 0.1608 \\

LGTSM~\cite{chang2019learnable} & 29.74 & 0.9504 & 0.070 & 0.1859 & 28.57 & 0.9409 & 0.170 & 0.1640 \\

CAP~\cite{lee2019copy} & 31.58 & 0.9607 & 0.071 & 0.1470 & 30.28 & 0.9521 & 0.182 & 0.1533 \\

FGVC~\cite{gao2020flow} & 29.67 & 0.9403 & 0.064 & 0.1022 & 30.80 & 0.9497 & 0.165 & 0.1586 \\

STTN~\cite{zeng2020learning} & 32.34 & 0.9655 & 0.053 & 0.0907 & 30.67 & 0.9560 & 0.149 & 0.1449 \\

FuseFormer~\cite{liu2021fuseformer} & 33.29 & 0.9681 & 0.053 & 0.0900 & 32.54 & 0.9700 & 0.138 & 0.1362 \\

E2FGVI~\cite{li2022towards} & 33.71 & 0.9700 & \underline{0.046} & \underline{0.0864} & 33.01 & 0.9721 & \textbf{0.116} & \underline{0.1315} \\

\hline  %
FlowLens & \underline{33.89} & \underline{0.9722} & \textbf{0.045} & 0.0869 & \underline{33.12} & \underline{0.9739} & \underline{0.120} & 0.1322 \\

\rowcolor{gray!20}
FlowLens+ & \textbf{34.23} & \textbf{0.9731} & 0.049 & \textbf{0.0843} & \textbf{33.47} & \textbf{0.9750} & 0.127 & \textbf{0.1282} \\

\hline

\end{tabular}
}
}

%% file: tables/revise_jetson_R2.tex
\resizebox{1.0\columnwidth}{!}{
\setlength{\tabcolsep}{3mm}{ %
\begin{tabular}{l|ccccc}

\hline
\multirow{2}{*}{\textbf{Method}} & \multirow{2}{*}{\textbf{PSNR$\uparrow$}} & \multirow{2}{*}{\textbf{SSIM$\uparrow$}} & \multirow{2}{*}{\textbf{VFID$\downarrow$}} & \multirow{2}{*}{\textbf{FLOPs}} & \textbf{Runtime} \\
& & & & & (s/frame) \\

\hline
\hline

{E2FGVI~\cite{li2022towards}} & 18.74 & 92.85 & 0.314 & 560G & 0.050\\ %

{FlowLens-s} & \textbf{19.04} & \textbf{93.22} & \textbf{0.308} & \textbf{207G} & \textbf{0.037}\\ %

\hline

\end{tabular}
}
}

%% file: tables/beyond-semantics-pinhole.tex
\resizebox{1.0\columnwidth}{!}{
\setlength{\tabcolsep}{0.5mm}{ %
\begin{tabular}{l|lll}

\hline
\multirow{2}{*}{\textbf{Method}} & \multicolumn{3}{c}{\textbf{mIoU($\%$)}}\\
& \textbf{Seen} & \textbf{Unseen} & \textbf{Overall}\\

\hline
\hline
FoV-limited Frames & 80.12 & 26.05 & 59.33\\

%
STTN~\cite{zeng2020learning} & 79.00$_{(-1.12)}$ & 38.66$_{(+12.61)}$ & 63.39$_{(+4.06)}$\\

E2FGVI~\cite{li2022towards} & 80.53$_{(+0.41)}$ & 40.89$_{(+14.84)}$ & 66.46$_{(+7.13)}$\\

\rowcolor{gray!20}
FlowLens & \textbf{82.42}$_{(+\bm{2.30})}$ & \textbf{45.04}$_{(+\bm{18.99})}$ & \textbf{69.29}$_{(+\bm{9.96})}$\\

\hline

\end{tabular}
}
}

%% file: tables/beyond-detection-pinhole.tex
\resizebox{1.0\columnwidth}{!}{
\setlength{\tabcolsep}{0.5mm}{ %
\begin{tabular}{l|llll}

\hline
\textbf{Method} & {$\text{AR}^{\text{box}}$} & {$\text{AP}^{\text{box}}$} & {$\text{AP}^{\text{box}}_{50}$} & {$\text{AP}^{\text{box}}_{75}$}\\

\hline
\hline
FoV-limited Frames & 36.7 & 34.1 & 36.6 & 34.4\\

%
STTN~\cite{zeng2020learning} & 40.3$_{(+3.6)}$ & 36.9$_{(+2.8)}$ & 39.6$_{(+3.0)}$ & 38.0$_{(+3.6)}$\\

E2FGVI~\cite{li2022towards} & 40.5$_{(+3.8)}$ & 37.3$_{(+3.2)}$ & 39.7$_{(+3.1)}$ & 38.6$_{(+4.2)}$\\

\rowcolor{gray!20}
FlowLens & \textbf{44.1}$_{(+\bm{7.4})}$ & \textbf{40.6}$_{(+\bm{6.5})}$ & \textbf{43.3}$_{(+\bm{6.7})}$ & \textbf{42.3}$_{(+\bm{7.9})}$\\

\hline

\end{tabular}
}
}

%% file: tables/beyond-semantics-sphere.tex
\resizebox{1.0\columnwidth}{!}{
\setlength{\tabcolsep}{0.5mm}{ %
\begin{tabular}{l|lll}

\hline
\multirow{2}{*}{\textbf{Method}} & \multicolumn{3}{c}{\textbf{mIoU($\%$)}}\\
& \textbf{Seen} & \textbf{Unseen} & \textbf{Overall}\\

\hline
\hline
FoV-limited Frames & 69.83 & 25.99 & 62.88\\

%
STTN~\cite{zeng2020learning} & 72.93$_{(+3.10)}$ & 34.69$_{(+8.70)}$ & 66.97$_{(+4.09)}$\\

E2FGVI~\cite{li2022towards} & 73.20$_{(+3.37)}$ & 36.43$_{(+10.44)}$ & 68.20$_{(+5.32)}$\\

\rowcolor{gray!20}
FlowLens & \textbf{73.93}$_{(+\bm{4.10})}$ & \textbf{39.62}$_{(+\bm{13.63})}$ & \textbf{69.71}$_{(+\bm{6.83})}$\\

\hline

\end{tabular}
}
}

%% file: tables/ablation-recurrent-v2.tex
\resizebox{1.0\columnwidth}{!}{
\setlength{\tabcolsep}{5mm}{ %
\begin{tabular}{l|cc}

\hline
\textbf{Aug. Layers} & \textbf{PSNR$\uparrow$} & \textbf{SSIM$\uparrow$} \\

\hline
\hline
$w/o$ & 36.08 & 0.9909 \\

\rowcolor{gray!20}
early & \textbf{36.49} & \textbf{0.9915} \\

middle & 36.13 & 0.9911 \\

%
late & 36.27 & 0.9912 \\

all & 36.32 & 0.9914 \\

%

\hline

\end{tabular}
}
}

%% file: tables/ablation-cross-v2.tex
\resizebox{1.0\columnwidth}{!}{
\setlength{\tabcolsep}{3mm}{ %
\begin{tabular}{l|ccc}

\hline
\textbf{Case} & \textbf{PSNR$\uparrow$} & \textbf{SSIM$\uparrow$} & \textbf{FLOPs}\\

\hline
\hline
Vanilla Attention~\cite{dosovitskiy2020image} & 35.98 & 0.9905 & 586.61G\\

%
Local Attention~\cite{liu2021swin} & 36.09 & 0.9913 & \textbf{584.00G} \\

%
Axial Attention~\cite{ho2019axial} & 36.24 & 0.9912 & 584.59G\\

%
Focal Attention~\cite{yang2021focal} & 36.15 & \textbf{0.9915} & 585.20G \\

\rowcolor{gray!20}
3D-Decoupled & \textbf{36.49} & \textbf{0.9915} & 586.64G \\

\hline

\end{tabular}
}
}

%% file: tables/revise_propainter.tex
\resizebox{1.0\columnwidth}{!}{
\setlength{\tabcolsep}{3.5mm}{    %
\begin{tabular}{lc|cccc}

\hline
\multirow{2}{*}{\textbf{Model}} & \multirow{2}{*}{\textbf{Extended FoV}} & \multirow{2}{*}{\textbf{PSNR$\uparrow$}} & \multirow{2}{*}{\textbf{SSIM$\uparrow$}} & \multirow{2}{*}{\textbf{VFID$\downarrow$}} & \textbf{Runtime$\downarrow$}\\
 &  &  &  &  & \textbf{(s/frame)}\\

\hline
\hline
\multirow{4}{*}{ProPainter~\cite{zhou2023propainter}} & 5\% & 24.19 & 98.26 & 0.186 & \multirow{4}{*}{\textbf{0.136}} \\

& 10\% & 21.03 & 95.46 & 0.260 & \\

& 20\% & 18.34 & 89.45 & 0.368 & \\

& \cellcolor[gray]{0.95}Avg. & \cellcolor[gray]{0.95}21.19 & \cellcolor[gray]{0.95}94.39 & \cellcolor[gray]{0.95}0.271 & \cellcolor[gray]{0.95} \\

\hline
\multirow{4}{*}{FlowLens (ProPainter)} & 5\% & 24.72 & 98.45 & 0.171 & \multirow{4}{*}{0.141}  \\

& 10\% & 21.53 & 95.92 & 0.239 &  \\

& 20\% & 18.56 & 90.31 & 0.354 &  \\

& \cellcolor[gray]{0.95}Avg. & \cellcolor[gray]{0.95}\textbf{21.60} & \cellcolor[gray]{0.95}\textbf{94.89} & \cellcolor[gray]{0.95}\textbf{0.255} & \cellcolor[gray]{0.95}  \\

\hline

\end{tabular}
}
}

%% file: tables/ablations-ddca.tex
\resizebox{1.0\columnwidth}{!}{
\setlength{\tabcolsep}{2.5mm}{ %
\begin{tabular}{cc|cc}

\hline
\textbf{Decoupling} & \textbf{Stip Pooling} & \textbf{PSNR$\uparrow$} & \textbf{SSIM$\uparrow$} \\

\hline
\hline
 \XSolidBrush & \XSolidBrush & 35.98 & 0.9905 \\

\Checkmark & \XSolidBrush & 36.29 & 0.9913 \\

\cellcolor{gray!20}\Checkmark & \cellcolor{gray!20}\Checkmark & \cellcolor{gray!20}\textbf{36.49} & \cellcolor{gray!20}\textbf{0.9915} \\

\hline

\end{tabular}
}
}

%% file: tables/ablation-flow.tex
\resizebox{1.0\columnwidth}{!}{
\setlength{\tabcolsep}{3.6mm}{ %
\begin{tabular}{l|cc}

\hline
\textbf{Case} & \textbf{PSNR$\uparrow$} & \textbf{SSIM$\uparrow$} \\

\hline
\hline
$w/o$ Flow Propagation & 35.31 & 0.9895\\

%
$w/o$ DCN Compensation & 35.16 & 0.9905\\

%
$w/$ SpyNet~\cite{ranjan2017optical} & 36.27 & \textbf{0.9915}\\

%
\rowcolor{gray!20}
$w/$ MaskFlowNetS~\cite{zhao2020maskflownet} & \textbf{36.49} & \textbf{0.9915}\\

\hline

\end{tabular}
}
}

%% file: tables/ablation-flow-res.tex
\resizebox{1.0\columnwidth}{!}{
\setlength{\tabcolsep}{1mm}{ %
\begin{tabular}{c|ccc}

\hline
 \textbf{Flow Resolution ($R$)} & \textbf{PSNR$\uparrow$} & \textbf{SSIM$\uparrow$} & \textbf{Runtime (s/frame)} \\

\hline
\hline
 1 & \textbf{36.54} & \underline{0.9914} & 0.055\\

$1/2$ & 36.41 & 0.9912 & 0.051\\ %

\cellcolor{gray!20}$1/4$ & \cellcolor{gray!20}\underline{36.49} & \cellcolor{gray!20}\textbf{0.9915} & \cellcolor{gray!20}\underline{0.049} \\

$1/8$ & 35.87 & 0.9702 & \textbf{0.048}\\

\hline

\end{tabular}
}
}

%% file: tables/ablation-mix.tex
\resizebox{1.0\columnwidth}{!}{
\setlength{\tabcolsep}{3.5mm}{    %
\begin{tabular}{l|cc}

\hline
\textbf{Case} & \textbf{PSNR$\uparrow$} & \textbf{SSIM$\uparrow$} \\

\hline
\hline
FlowLens $w/$ FFN~\cite{dosovitskiy2020image} & 36.14 & 0.9910 \\

%
FlowLens $w/$ F3N~\cite{liu2021fuseformer} & 36.15 & 0.9912 \\

%
FlowLens $w/$ Mix-FFN~\cite{xie2021segformer} & 36.30 & \textbf{0.9915} \\

\rowcolor{gray!20}
FlowLens $w/$ MixF3N & \textbf{36.49} & \textbf{0.9915} \\

\hline

\end{tabular}
}
}

%% file: supp.tex
\clearpage
\appendices
\counterwithin{figure}{section}
\counterwithin{equation}{section}


\section{Evaluation Details}

\begin{figure*}
    \centering
    \includegraphics[width=1\textwidth]{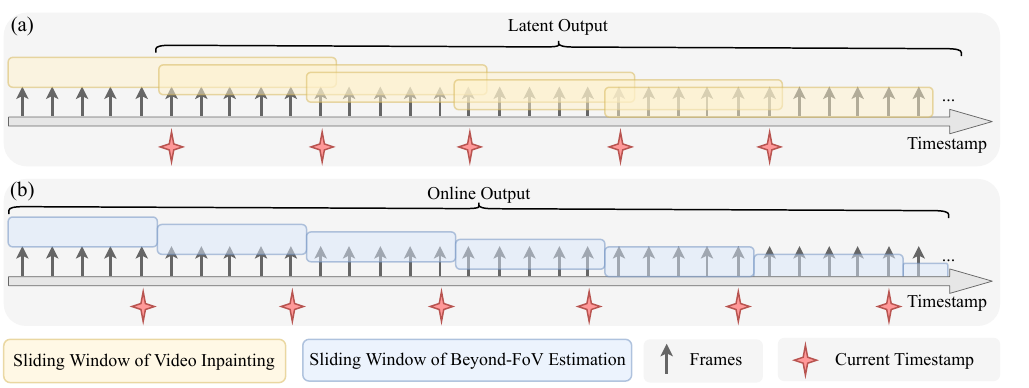}
    \caption{The evaluation logic comparison of offline video inpainting and online video inpainting of beyond-FoV scene reconstruction. (a) Video inpainting is regarded as a video editing technique and adopts overlapping sliding windows to get latent output. (b) Future frames cannot be accessed in the beyond-FoV estimation, thus the sliding windows are non-overlapping and output online.}
    \vspace{1.0em}
    \label{fig:sup_test_logic}
\end{figure*}

Following~\cite{li2022towards}, we test our models with the resolution of $432{\times}240$ on YouTube-VOS~\cite{xu2018youtube} and DAVIS~\cite{perazzi2016benchmark} with the same test mask. For the offline video inpainting, we adopt the same evaluation pipeline of the previous work~\cite{li2022towards} that uses the sliding window with a size of $10$ and sample the reference frame from the entire video with a stride of $10$. For online video inpainting of beyond-FoV scene reconstruction, we consider that only past reference frames can be used, and the sliding window size is set to $5$ with $3$ past reference frames. Note that the sliding window ends with $t{+}5$ for the offline video inpainting but $t$ for the online at time $t$, since the future information cannot be accessed in this track (see Fig.~\ref{fig:sup_test_logic}). The FLOPs are computed using input images of $432{\times}240$ of temporal length $8$. The speed test is 
performed on a single RTX3090 GPU.

\section{Additional Discussion}
\subsection{Detailed Results on Different Expansion Rate}
We report the average results for different FoV expansions to save pages in the main text. Therefore, in Tab.~\ref{tab:fov_expansion_supp}, we further show the detailed performance of alternative approaches on various FoV expansion rates. We conclude that the proposed FlowLens/-s outperforms previous works across different fields of view.

\begin{table*}[!t]
    \begin{center}
        \vspace{3.0em}
        \caption{\emph{Quantitative comparisons on KITTI360 beyond-FoV scene reconstruction.} Inner FoV expansion results are not reported for SRN, which is an image outpainting method. $E_{warp}^{*}$ denotes $E_{warp} {\times} 10^{-2}$. The best are shown in \textbf{bold}, and the second best are \underline{underlined}.}
        \label{tab:fov_expansion_supp}
        \input{tables/fov-expansion-v2-gray}
    \end{center}
\end{table*}

\subsection{Can We Only Use Past Frames as Input Clip?}
In this section, we further discuss the online output capability of FlowLens for FoV expansion. In the default inference behavior, FlowLens only samples reference frames from the past, and each frame is inferred only once, which is defined as ``online mode''. To simultaneously complete all the input frames in a single feed-forward process, we follow the popular ``multi-to-multi'' paradigm~\cite{liu2021fuseformer,li2022towards,zeng2020learning}. As shown in Tab.~\ref{tab:reb_window_stride}, FlowLens can achieve a better accuracy-speed trade-off than previous VI-Trans on both offline and online mode even without leaking any local future information when window\&stride=1, \ie, only clips consisting of past frames are fed into the model to expand the FoV.

\begin{table}[h]
    \begin{center}
        \caption{\emph{Ablation study of different sliding window strategies on KITTI360 inner FoV expansion set.} Future Sample means sample reference frame from future.}
        \label{tab:reb_window_stride}
        \input{tables/reb_window_stridev3}
    \end{center}
\end{table}

\subsection{How to Choose the Local Window Size?}
The local window size matters for the full dissemination and interaction of fine-grained information in a video, while the computational overhead will also increase accordingly. However, we did not find any ablation experiments for this hyper-parameter in previous VI-Trans work~\cite{liu2021fuseformer,li2022towards,zeng2020learning}. Therefore, we conduct ablations on local window size in Tab.~\ref{tab:reb_window_size} and set it to $5$ for holding a fine accuracy-efficiency trade-off. 

\begin{table}[h]
    \begin{center}
        \caption{\emph{Effect of local window size on inner FoV expansion set.}}
        \label{tab:reb_window_size}
        \vspace{-1.0em}
        \input{tables/reb_window_sizev2}
        \vspace{-2.5em}
    \end{center}
\end{table}

\subsection{Does a Longer Memory Help?}
We conduct ablations on $w/o$ Clip-Recurrent Hub as shown in Tab.~\ref{tab:reb_clip_hub}.
We store key-value pairs of the previous clip instead of fused clips to avoid error accumulation and amplify in the next step~\cite{kang2022ecfvi}.
We experiment with propagating fused clips and it causes an accuracy drop. We also explore the linear layers to aggregate key-value pairs of different stamps to extend the cached clip time length like MeMViT~\cite{wu2022memvit},
but no significant accuracy gain is observed, which can be attributed to that different temporal pixel distributions are mixed and thus interfere with the current query operation as beyond-FoV estimation is a fine-grained per-pixel task, and every clean pixel matters.

\begin{table}[t]
    \begin{center}
        \caption{\emph{Effect of stored key-value pairs in the clip-recurrent hub.}}
        \label{tab:reb_clip_hub}
        \input{tables/reb_clip_hubv2}
    \end{center}
\end{table}

\section{Potential negative impact}
In this paper, we propose FlowLens, which aims to see the world outside the physical FoV of information-collecting sensors. However, if this technique is used by people with ulterior motives for military purposes, such as battlefield perception systems on armored vehicles, terminal visual guidance systems for missiles, \etc, it may exacerbate military conflicts in disputed areas and affect world peace and stability. In the meantime, the results of expanding the FoV outward have not been perfect, which may become a safety hazard in applications such as autonomous driving and mobile robots. In the future, We will continue to explore beyond-FoV estimation to further improve the reliability and robustness of the algorithm.

%% file: tables/fov-expansion-v2-gray.tex
\resizebox{\textwidth}{!}{ 
\setlength{\tabcolsep}{2mm}{ %
\begin{tabular}{c|c|c|cccc:cccc|c|c}

\hline
\multicolumn{2}{c|}{\multirow{3}{*}{\textbf{Method}}} & \multirow{3}{*}{\textbf{Ex. Rate}} & \multicolumn{8}{c|}{\textbf{KITTI360}} & \multicolumn{2}{c}{\textbf{Efficiency}} \\
\cline{4-13}
\multicolumn{1}{c}{}&&& \multicolumn{4}{c:}{Outer FoV Expansion} & \multicolumn{4}{c|}{Inner FoV Expansion} & \multirow{2}{*}{FLOPs} & \multirow{2}{*}{\makecell[c]{Runtime \\ (s/frame)}}\\ %

\cline{4-11}
\multicolumn{1}{c}{}&&& PSNR$\uparrow$ & SSIM$\uparrow$ & VFID$\downarrow$ & $E_{warp}^{*}\downarrow$ & PSNR$\uparrow$ & SSIM$\uparrow$ & VFID$\downarrow$ & $E_{warp}^{*}\downarrow$ & &\\

\hline
\multirow{9}{*}{Image Inpainting} & \multirow{4}{*}{LaMa~\cite{suvorov2022resolution}} & $5\%$ & 21.47 & 0.9615 & 0.279 & 0.5157 & 34.41 & 0.9946 & 0.015 & 0.3570 & \multirow{4}{*}{778.4G} & \multirow{4}{*}{0.088} \\
\cline{3-11}
&& $10\%$ & 18.76 & 0.9183 & 0.372 & 0.5831 & 32.62 & 0.9918 & 0.020 & 0.3655 &  &  \\
\cline{3-11}
&& $20\%$ & 16.70 & 0.8420 & 0.414 & 0.6419 & 28.03 & 0.9764 & 0.043 & 0.3961 &  &  \\
\cline{3-11}
& & \cellcolor{gray!20}$Avg.$ & \cellcolor{gray!20}18.98 & \cellcolor{gray!20}0.9073 & \cellcolor{gray!20}0.355 & \cellcolor{gray!20}0.5802 & \cellcolor{gray!20}31.69 & \cellcolor{gray!20}0.9876 & \cellcolor{gray!20}0.026 & \cellcolor{gray!20}0.3729 &  &  \\

\cline{2-13}
\multirow{4}{*}{$\&$}& \multirow{4}{*}{PUT~\cite{liu2022reduce}} & $5\%$ & 21.71 & 0.9704 & 0.232 & 0.5490 & 27.34 & 0.9743 & 0.071 & 0.5035 & \multirow{4}{*}{-} & \multirow{4}{*}{14.83} \\
\cline{3-11}
&& $10\%$ & 18.29 & 0.9181 & 0.329 & 0.6549 & 26.80 & 0.9710 & 0.091 & 0.5174 &  &  \\
\cline{3-11}
&& $20\%$ & 16.10 & 0.8176 & 0.403 & 1.1308 & 24.65 & 0.9519 & 0.142 & 0.5770 &  &  \\
\cline{3-11}
\multirow{1}{*}{Outpainting}& & \cellcolor{gray!20}$Avg.$ & \cellcolor{gray!20}18.70 & \cellcolor{gray!20}0.9020 & \cellcolor{gray!20}0.321 & \cellcolor{gray!20}0.7782 & \cellcolor{gray!20}26.26 & \cellcolor{gray!20}0.9657 & \cellcolor{gray!20}0.101 & \cellcolor{gray!20}0.5326 &  &  \\

\cline{2-13}
&\multirow{4}{*}{SRN~\cite{wang2019wide}} & $5\%$ & 17.72 & 0.9283 & 0.418 & 0.5391 & - & - & - & - & \multirow{4}{*}{-} & \multirow{4}{*}{0.032} \\
\cline{3-11}
&& $10\%$ & 15.85 & 0.8510 & 0.490 & 0.5996 & - & - & - & - &  & \\
\cline{3-11}
&& $20\%$ & 14.74 & 0.7523 & 0.528 & 0.6808 & - & - & - & - &  &  \\
\cline{3-11}
& & \cellcolor{gray!20}$Avg.$ & \cellcolor{gray!20}16.10 & \cellcolor{gray!20}0.8439 & \cellcolor{gray!20}0.479 & \cellcolor{gray!20}0.6065 & \cellcolor{gray!20}- & \cellcolor{gray!20}- & \cellcolor{gray!20}- & \cellcolor{gray!20}- &  &  \\

\hline
\hline

\multirow{12}{*}{Video Inpainting} & \multirow{4}{*}{STTN~\cite{zeng2020learning}} & $5\%$ & 21.34 & 0.9685 & 0.269 & 0.5248 & 42.46 & 0.9989 & 0.006 & 0.3623 & \multirow{4}{*}{885.6G} & \multirow{4}{*}{0.019} \\
\cline{3-11}
&& $10\%$ & 18.73 & 0.9232 & 0.342 & 0.6208 & 36.26 & 0.9957 & 0.015 & 0.3701 &  & \\
\cline{3-11}
&& $20\%$ & 16.13 & 0.8317 & 0.501 & 0.8154 & 28.80 & 0.9783 & 0.048 & 0.3864 &  & \\
\cline{3-11}
& & \cellcolor{gray!20}$Avg.$ & \cellcolor{gray!20}18.73 & \cellcolor{gray!20}0.9078 & \cellcolor{gray!20}0.371 & \cellcolor{gray!20}0.6537 & \cellcolor{gray!20}35.84 & \cellcolor{gray!20}0.9909 & \cellcolor{gray!20}\underline{0.023} & \cellcolor{gray!20}0.3729 &  & \\

\cline{2-13}
& \multirow{4}{*}{FuseFormer~\cite{liu2021fuseformer}} & $5\%$ & 21.68 & 0.9707 & 0.259 & 0.5523 & 41.83 & 0.9986 & 0.004 & 0.3650 & \multirow{4}{*}{446.9G} & \multirow{4}{*}{0.033} \\
\cline{3-11}
&& $10\%$ & 18.91 & 0.9266 & 0.368 & 0.6775 & 35.07 & 0.9942 & 0.019 & 0.3771 &  & \\
\cline{3-11}
&& $20\%$ & 16.14 & 0.8389 & 0.514 & 0.8901 & 27.45 & 0.9712 & 0.078 & 0.4102 &  & \\
\cline{3-11}
& & \cellcolor{gray!20}$Avg.$ & \cellcolor{gray!20}18.91 & \cellcolor{gray!20}0.9121 & \cellcolor{gray!20}0.380 & \cellcolor{gray!20}0.7066 & \cellcolor{gray!20}34.78 & \cellcolor{gray!20}0.9880 & \cellcolor{gray!20}0.034 & \cellcolor{gray!20}0.3841 &  & \\

\cline{2-13}
& \multirow{4}{*}{E2FGVI~\cite{li2022towards}} & $5\%$ & 22.32 & 0.9746 & 0.213 & 0.4778 & 42.73 & 0.9989 & 0.004 & 0.3618 & \multirow{4}{*}{560.0G} & \multirow{4}{*}{0.038} \\
\cline{3-11}
&& $10\%$ & 19.32 & 0.9354 & 0.315 & 0.5635 & 36.43 & 0.9957 & 0.021 & 0.3694 &  & \\
\cline{3-11}
&& $20\%$ & 16.71 & 0.8588 & 0.412 & 0.7034 & 28.56 & 0.9767 & 0.061 & 0.3871 &  & \\
\cline{3-11}
& & \cellcolor{gray!20}$Avg.$ & \cellcolor{gray!20}19.45 & \cellcolor{gray!20}0.9229 & \cellcolor{gray!20}0.313 & \cellcolor{gray!20}0.5816 & \cellcolor{gray!20}35.91 & \cellcolor{gray!20}0.9904 & \cellcolor{gray!20}0.029 & \cellcolor{gray!20}0.3728 &  & \\

\hline
\hline

\multirow{12}{*}{\textbf{Ours}} & \multirow{4}{*}{\textbf{FlowLens}} & $5\%$ & 23.07 & 0.9781 & 0.185 & 0.4759 & 43.99 & 0.9992 & 0.004 & 0.3600 & \multirow{4}{*}{586.6G} & \multirow{4}{*}{0.049} \\
\cline{3-11}
&& $10\%$ & 20.00 & 0.9427 & 0.282 & 0.5175 & 37.03 & 0.9963 & 0.018 & 0.3627 &  & \\
\cline{3-11}
&& $20\%$ & 17.32 & 0.8733 & 0.375 & 0.6107 & 29.06 & 0.9793 & 0.058 & 0.3790 &  & \\
\cline{3-11}
&& \cellcolor{gray!20}$Avg.$ & \cellcolor{gray!20}\underline{20.13} & \cellcolor{gray!20}\underline{0.9314} & \cellcolor{gray!20}\textbf{0.281} & \cellcolor{gray!20}0.5347 & \cellcolor{gray!20}\underline{36.69} & \cellcolor{gray!20}\underline{0.9916} & \cellcolor{gray!20}0.027 & \cellcolor{gray!20}0.3672 &  & \\

\cline{2-13}    %
& \multirow{4}{*}{\textbf{FlowLens-s}} & $5\%$ & 22.35 & 0.9743 & 0.209 & 0.4860 & 42.71 & 0.9989 & 0.008 & 0.3609 & \multirow{4}{*}{207.2G} & \multirow{4}{*}{0.023} \\
\cline{3-11}
&& $10\%$ & 19.56 & 0.9365 & 0.295 & 0.4990 & 36.56 & 0.9959 & 0.021 & 0.3644 &  & \\
\cline{3-11}
&& $20\%$ & 17.13 & 0.8634 & 0.395 & 0.5401 & 29.25 & 0.9800 & 0.062 & 0.3726 &  & \\
\cline{3-11}
&& \cellcolor{gray!20}$Avg.$ & \cellcolor{gray!20}19.68 & \cellcolor{gray!20}0.9247 & \cellcolor{gray!20}\underline{0.300} & \cellcolor{gray!20}\underline{0.5084} & \cellcolor{gray!20}36.17 & \cellcolor{gray!20}\underline{0.9916} & \cellcolor{gray!20}0.030 & \cellcolor{gray!20}\underline{0.3660} &  & \\

\cline{2-13}    %
& \multirow{4}{*}{\textbf{FlowLens+}} & $5\%$ & 23.51 & 0.9794 & 0.194 & 0.3908 & 44.58 & 0.9993 & 0.003 & 0.3588 & \multirow{4}{*}{586.6G} & \multirow{4}{*}{0.148} \\
\cline{3-11}
&& $10\%$ & 20.29 & 0.9438 & 0.297 & 0.3868 & 37.72 & 0.9968 & 0.014 & 0.3590 &  & \\
\cline{3-11}
&& $20\%$ & 17.70 & 0.8733 & 0.409 & 0.3963 & 29.83 & 0.9817 & 0.041 & 0.3588 &  & \\
\cline{3-11}
&& \cellcolor{gray!20}$Avg.$ & \cellcolor{gray!20}\textbf{20.50} & \cellcolor{gray!20}\textbf{0.9322} & \cellcolor{gray!20}\underline{0.300} & \cellcolor{gray!20}\textbf{0.3913} & \cellcolor{gray!20}\textbf{37.38} & \cellcolor{gray!20}\textbf{0.9926} & \cellcolor{gray!20}\textbf{0.019} & \cellcolor{gray!20}\textbf{0.3589} &  & \\

\hline

\end{tabular}
}
}

%% file: tables/reb_window_stridev3.tex
\resizebox{1.0\columnwidth}{!}{
\setlength{\tabcolsep}{1mm}{ %
\begin{tabular}{l|cccc|ccc}

\hline
\textbf{Method} & \textbf{Online} & \textbf{Window} & \textbf{Stride} & \textbf{Future Sample} & \textbf{PSNR$\uparrow$} & \textbf{SSIM$\uparrow$} & \textbf{FLOPs}\\

\hline
\hline
\multirow{3}{*}{E2FGVI} & \XSolidBrush & 11 & 5 & \Checkmark & 36.69 & 0.9918 & 1445.0G \\

& \Checkmark & 5 & 5 & \XSolidBrush & 35.91 & 0.9904 & 560.0G \\

& \Checkmark & 1 & 1 & \XSolidBrush & 33.89 & 0.9861 & 230.4G \\

\hline
\multirow{3}{*}{FlowLens-S} & \XSolidBrush & 11 & 5 & \Checkmark & \textbf{37.07} & \textbf{0.9932} & 534.9G \\

& \cellcolor{gray!20}\Checkmark & \cellcolor{gray!20}5 & \cellcolor{gray!20}5 & \cellcolor{gray!20}\XSolidBrush & \cellcolor{gray!20}36.17 & \cellcolor{gray!20}0.9916 & \cellcolor{gray!20}207.2G \\

& \Checkmark & 1 & 1 & \XSolidBrush & 34.30 & 0.9875 & \textbf{80.8G} \\

\hline

\end{tabular}
}
}

%% file: tables/reb_window_sizev2.tex
\resizebox{1.0\columnwidth}{!}{
\setlength{\tabcolsep}{2mm}{ %
\begin{tabular}{l|c|ccc}

\hline
\textbf{Method} & \textbf{LF Window Size} & \textbf{PSNR$\uparrow$} & \textbf{SSIM$\uparrow$} & \textbf{FLOPs} \\

\hline
\hline
\multirow{3}{*}{FlowLens} & 3 & 35.85 & 0.9903 & \textbf{404.2G}\\

 & \cellcolor{gray!20}5 & \cellcolor{gray!20}\underline{36.49} & \cellcolor{gray!20}\underline{0.9915} & \cellcolor{gray!20}\underline{586.6G}\\

 & 7 & \textbf{36.54} & \textbf{0.9917} & 734.7G\\

\hline

\end{tabular}
}
}

%% file: tables/reb_clip_hubv2.tex
\resizebox{1.0\columnwidth}{!}{
\setlength{\tabcolsep}{1mm}{ %
\begin{tabular}{c|ccc|cc}

\hline
\textbf{Method} & \textbf{Clip-Recurrent Hub} & \textbf{Key-Value Pairs} & \textbf{Mem Len.} & \textbf{PSNR$\uparrow$} & \textbf{SSIM$\uparrow$} \\

\hline
\hline
\multirow{5}{*}{FlowLens} & \XSolidBrush & w/o hub & w/o hub & 36.08 & 0.9909 \\

 & \Checkmark & pre. fusion & 1 & 36.16 & 0.9909 \\

 & \cellcolor{gray!20}\Checkmark & \cellcolor{gray!20}pre. clip & \cellcolor{gray!20}1 & \cellcolor{gray!20}\textbf{36.49} & \cellcolor{gray!20}\textbf{0.9915} \\

 & \Checkmark & pre. clip & 2 & 36.23 & 0.9910 \\

 & \Checkmark & pre. clip & 4 & 36.28 & 0.9913 \\

\hline

\end{tabular}
}
}